%% file: main_camera_ready_compact_for_arxiv.tex
\documentclass[10pt,twocolumn,letterpaper]{article}
\usepackage[accsupp]{axessibility}

\usepackage[pagenumbers]{cvpr} 

\input{preamble}
\definecolor{cvprblue}{rgb}{0.21,0.49,0.74}
\usepackage[pagebackref,breaklinks,colorlinks,allcolors=cvprblue]
{hyperref}

\usepackage{booktabs}
\usepackage{multirow}
\usepackage[table,xcdraw]{xcolor}
\definecolor{lightyellow}{RGB}{255, 255, 224}

\def\paperID{38678} 
\def\confName{CVPR}
\def\confYear{2026}

\title{VFM-VAE: Vision Foundation Models Can Be Good Tokenizers for Latent Diffusion Models}

\author{Tianci Bi$^1$\footnotemark[1] \quad Xiaoyi Zhang$^2$ \quad Yan Lu$^2$\footnotemark[2] \quad Nanning Zheng$^1$\footnotemark[2] \\
$^1$ State Key Laboratory of Human-Machine Hybrid Augmented Intelligence, \\
Institute of Artificial Intelligence and Robotics, Xi’an Jiaotong University \\
$^2$ Microsoft Research Asia \\
\tt \small tiancibi@stu.xjtu.edu.cn, \{xiaoyizhang, yanlu\}@microsoft.com, nnzheng@mail.xjtu.edu.cn}

\begin{document}
\maketitle
\footnotetext[1]{Work done during the internship at Microsoft Research Asia.}
\footnotetext[2]{Corresponding authors.}   
\input{sec_camera_ready/0_abstract}
\input{sec_camera_ready/1_intro}
\input{sec_camera_ready/2_related_work}

\input{sec_camera_ready/3_method_compact}
\input{sec_camera_ready/4_experiments_compact_for_arxiv}
\input{sec_camera_ready/5_conclusion}
\clearpage
\input{sec_camera_ready/6_acknowledgments}
{
    \small
    \bibliographystyle{ieeenat_fullname}
    \bibliography{main}
}

\input{sec_camera_ready/X_suppl}

\end{document}

%% file: sec_camera_ready/0_abstract.tex
\begin{abstract}
The performance of Latent Diffusion Models (LDMs) is critically dependent on the quality of their visual tokenizers. While recent works have explored incorporating Vision Foundation Models (VFMs) into the tokenizers training via distillation, we empirically find this approach inevitably weakens the robustness of learnt representation from original VFM.
In this paper, we bypass the distillation by proposing a more direct approach by leveraging the frozen VFM for the LDMs tokenizer, named VFM Variational Autoencoder (VFM-VAE).
To fully exploit the potential to leverage frozen VFM for the LDMs tokenizer, we design a new decoder to reconstruct realistic images from the semantic-rich representation of VFM.
With the proposed VFM-VAE, we conduct a systematic study on how the representation from different tokenizers impact the representation learning process throughout diffusion training, enabling 
synergistic benefits of dual-side alignment on both tokenizers and diffusion models.
Our effort in tokenizer design and training strategy lead to superior performance and efficiency: our system reaches a \textbf{gFID (w/o CFG) of 2.22 in merely 80 epochs} (a $10\times$ speedup over prior tokenizers). 
With continued training to \textbf{640 epochs}, it further attains a \textbf{gFID (w/o CFG) of 1.62}.
These results offer solid evidence for the substantial potential of VFMs to serve as visual tokenizers to accelerate the LDM training progress.
\end{abstract}

\begin{figure}[t]
  \centering
  \includegraphics[width=0.87\linewidth]{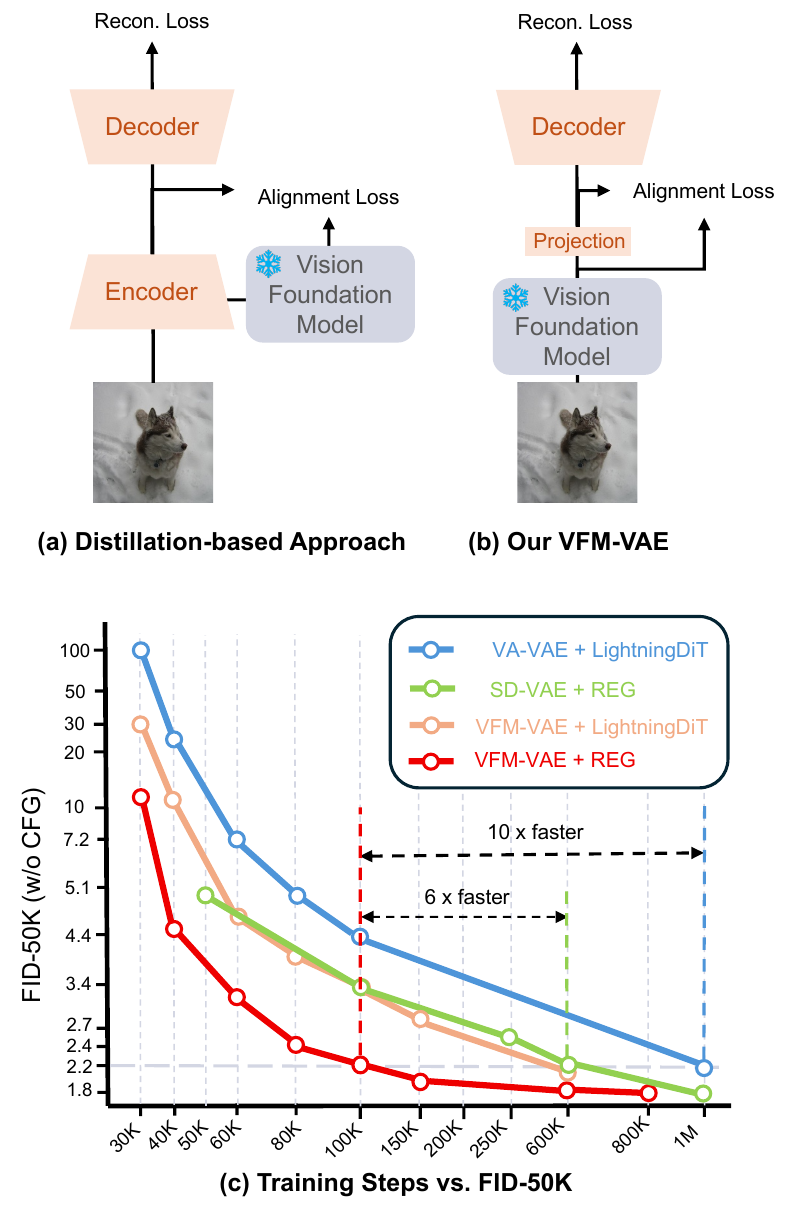}
   \caption{Comparison of VFM-VAE and previous visual tokenizers for LDM.
    (a) Distillation-based approach: VAE variants~\cite{vavae,repae} distill advanced representation from VFM.
    (b) Our VFM-VAE: directly leverage frozen VFM as the encoder front of the VAE.
    (c) Combining our visual tokenizer with various diffusion models leads to faster convergence and better performance.}
   \label{fig:abstract}
   \vspace{-1em}
\end{figure}

%% file: sec_camera_ready/1_intro.tex
\section{Introduction}
\label{sec:intro}

Latent Diffusion Models (LDMs)~\cite{ldm} have emerged as the dominant paradigm in visual synthesis, achieving impressive fidelity through an elegant two-stage framework: first train a visual tokenizer (typically a Variational Autoencoder~\cite{vae}) that encodes raw pixel-space images into a compact latent space, then learn the diffusion process within this representation space. This approach has proven remarkably effective in training scalable generative models while significantly reducing computational requirements.

\begin{figure*}[t]
  \centering
  \includegraphics[width=\linewidth]{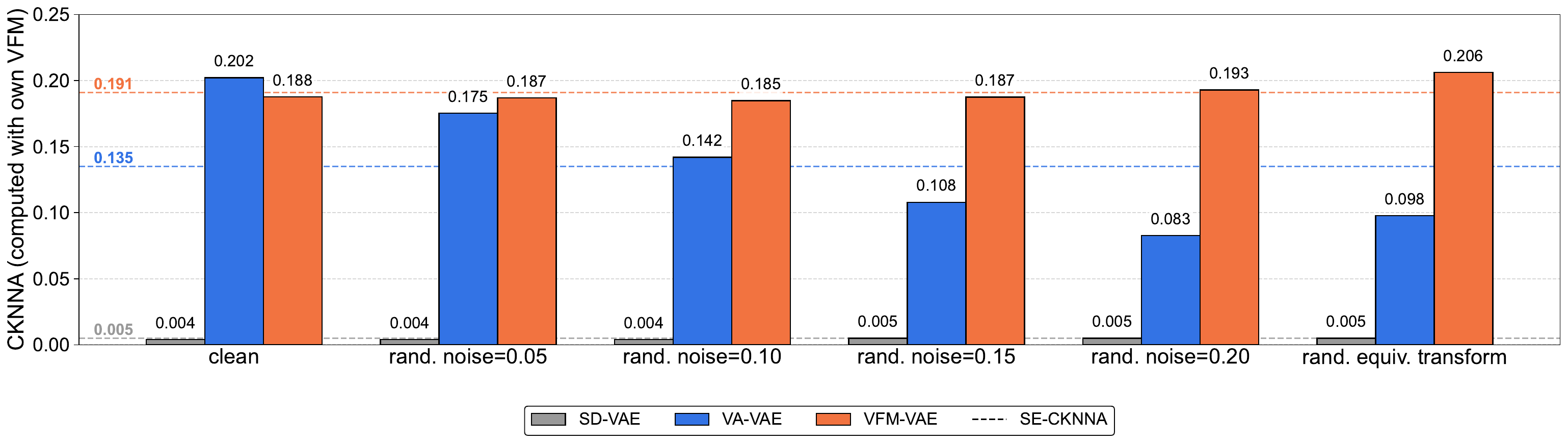}
   \caption{Brittleness of aligned representations under semantic-preserving transformations. Specifically, CKNNA~\cite{platonic} values for VA-VAE~\cite{vavae} and SD-VAE~\cite{ldm} are computed with DINOv2-Large~\cite{dinov2}, while those for VFM-VAE are computed with SigLIP2-Large~\cite{siglip2}. Under semantic-preserving transformations, VFM-VAE demonstrates notably stronger alignment with VFM features than VA-VAE.}
  \label{fig:brittleness}
  \vspace{-1em}
\end{figure*}

The quality of latent representations produced by the visual tokenizer is crucial to the success of the diffusion training process. Recently, numerous works~\cite{vavae,repae,imagefolder,detok} have explored incorporating Vision Foundation Model (VFM) representations into visual tokenizers, motivated by the significant progress in self-supervised~\cite{dinov2,dinov3} and weakly-supervised~\cite{clip,siglip,siglip2} representation learning. For example, VA-VAE~\cite{vavae} aligns VAE latents with VFM features through a carefully designed similarity loss, while REPA-E~\cite{repae} jointly trains VAE and the diffusion model to achieve alignment of VFM representations in the latter.


Despite these promising developments, we identify a fundamental limitation: \textbf{alignment-based distillation inevitably introduces representation degradation compared to the original VFM}. Measured by CKNNA~\cite{platonic}, a common metric for feature similarity, our analysis shows these representations exhibit unexpected brittleness under semantic-preserving transformations (\cref{fig:brittleness}), indicating critical information loss during distillation.

This observation motivates our key insight: rather than training a VAE to mimic VFM representations through distillation, we should directly utilize frozen VFM encoders within the VAE framework and built a pixel decoder for it. While conceptually straightforward, this approach faces a significant challenge: VFMs are optimized for semantic understanding rather than pixel-level reconstruction, creating a fundamental tension between semantic richness and reconstruction fidelity when paired with standard decoders.


Inspired by the success of VFMs in dense prediction tasks~\cite{llava, pencoder}, we hypothesize that frozen VFM encoders can enable high-fidelity image reconstruction with appropriate architectural adaptations. To this end, we enhance the standard VAE decoder with two key innovations: \textbf{Multi-Scale Latent Fusion} and \textbf{Progressive Resolution Reconstruction Blocks}. The \emph{Multi-Scale Latent Fusion} mechanism effectively leverages the hierarchical information in VFM features across different semantic levels, while the \emph{Progressive Resolution Reconstruction Blocks} enable stable training through advanced synthesis architectures and multi-resolution supervision. This design addresses the challenge of reconstructing pixel-accurate images from semantically-rich but spatially-coarse VFM representations.


Combining these architectural innovations with a objective that balances semantic preservation and reconstruction fidelity, we present \textbf{Vision Foundation Model Variational Autoencoder (VFM-VAE)}, the first framework to enable high-fidelity image reconstruction from a compact latent space, directly driven by frozen VFM encoders.


Building on VFM-VAE, we assess the representation alignment across visual tokenizers of LDMs. To this end, we extend CKNNA~\cite{platonic} to SE-CKNNA, which more reliably measures tokenizer and VFM alignment under perturbation. We analyze how different tokenizers impact diffusion representation learning and find two clear trends: (1) \textit{High-quality tokenizer representation facilitates diffusion representation learning}, which can be observed throughout all layers of diffusion models. (2) \textit{Dual-side alignment on both tokenizers and diffusion models yields more VFM-aligned representation inside diffusion models}, especially in deeper layers. Incorporating representation-aware supervision to strengthen these layers brings notable efficiency gains: our system reaches a gFID (w/o CFG) of 2.22 in 80 epochs, achieving a 10$\times$ speedup over previous tokenizers.

\begin{figure*}[t]
  \centering
  \includegraphics[width=\linewidth]{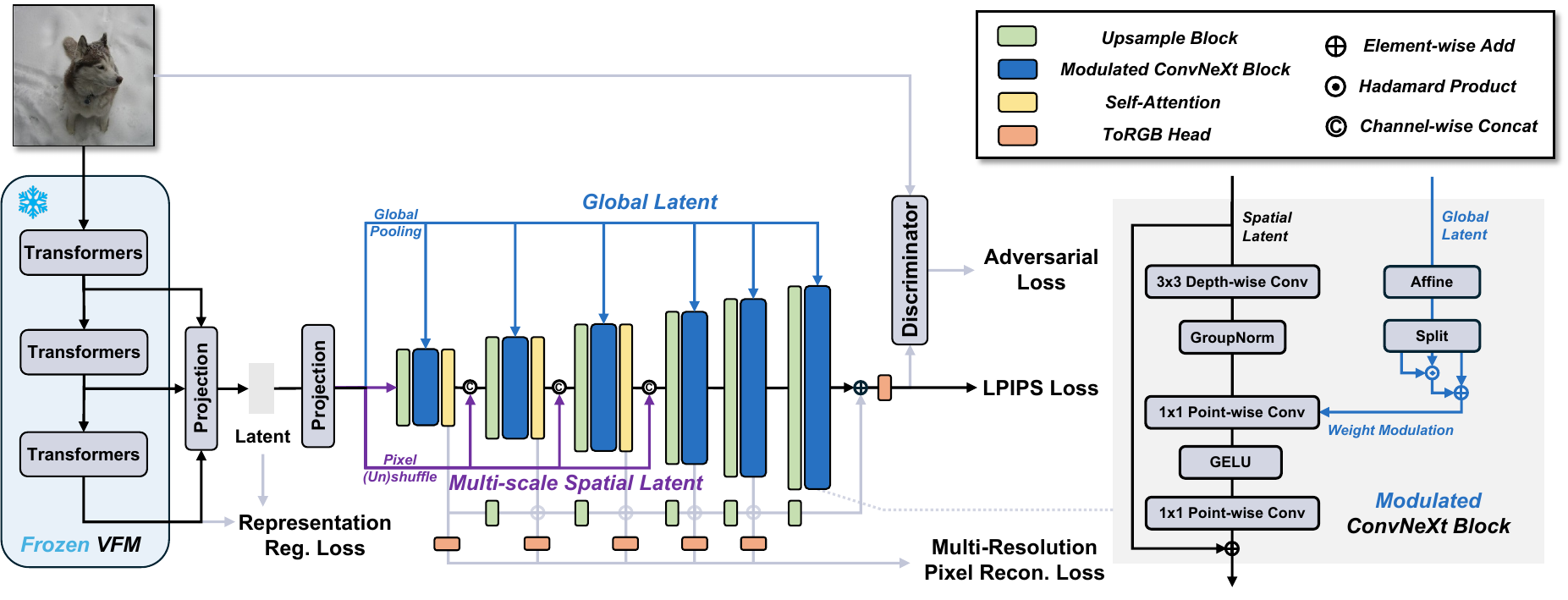}
   \caption{Overview of VFM-VAE architecture design. The model couples a frozen VFM encoder with a carefully designed multi-scale decoder to preserve semantic alignment and enable high-fidelity reconstruction even under highly compressed latent conditions.}
   \label{fig:vfmvae_arch}
\end{figure*}

In summary, our contributions are threefold:

\begin{itemize}
\item We propose to directly use a frozen VFM as its encoder front for latent diffusion, removing distillation-induced degradation while preserving high reconstruction quality through specialized decoder designs.
\item We introduce SE-CKNNA as a diagnostic metric for tokenizer–VFM alignment and analyze how tokenizer latents influence diffusion representation learning, which leads to a joint tokenizer–diffusion alignment strategy that more effectively leverages VFM representations.
\item We achieve solid results on ImageNet $256 \times 256$: with our alignment strategy, VFM-VAE reaches a gFID (w/o CFG) of 1.62 in 640 epochs, demonstrating superior performance and efficiency.
\end{itemize}

%% file: sec_camera_ready/2_related_work.tex
\section{Related work}
\label{sec:related_work}

\subsection{Visual tokenizers for latent diffusion}

Standard latent diffusion models typically rely on SD-VAE~\cite{ldm}, whose limited semantic capacity motivates recent efforts to incorporate Vision Foundation Models (VFMs). VA-VAE~\cite{vavae} aligns VAE latents with DINOv2~\cite{dinov2} using a similarity loss, while REPA-E~\cite{repae} jointly trains the tokenizer and diffusion model to improve semantic consistency. Other tokenizers based on VQGAN~\cite{vqgan} or folded-token designs~\cite{imagefolder} improve compression but often sacrifice reconstruction fidelity. All these methods distill representations from VFMs and thus suffer from degradation under semantic-preserving perturbations.

\subsection{Representation alignment in diffusion models}

Recent work has begun to explicitly align the representations of diffusion models with those of VFMs. REPA~\cite{repa} introduces semantic supervision from DINOv2~\cite{dinov2} to guide the intermediate features of diffusion transformers, while REG~\cite{reg} further extends this idea by incorporating class-token alignment to strengthen semantic consistency during generation. These methods show that VFM-guided alignment can substantially improve the representational quality of diffusion models; however, they operate under the assumption that the tokenizer already provides a stable and semantically reliable latent space. In contrast, we investigate how the tokenizer’s latent representations influence the evolution of diffusion features and propose architectural and training designs that preserve VFM semantics while simultaneously supporting high-fidelity reconstruction.

\subsection{The relation with concurrent works}

Unlike concurrent works (RAE~\cite{rae} and SVG~\cite{svg}) that directly use high-dimensional VFM features, we adhere to the latent compression of LDMs. This ensures VFM-VAE remains seamlessly compatible with existing diffusion frameworks. We also systematically explore how VFM-based encoders improve representation quality. A detailed discussion is provided in the Supplementary Material Sec.~1.

%% file: sec_camera_ready/3_method_compact.tex
\section{Method}
\label{sec:method}



Directly using a frozen VFM as the encoder of a generative tokenizer brings several non-trivial challenges. First, VFM feature hierarchies are \textbf{optimized for semantic understanding} rather than pixel-level reconstruction. Second, VFM feature maps have coarse spatial resolution and highly anisotropic channel distributions, making it \textbf{difficult for a standard decoder} to recover fine-grained details. Third, preserving VFM semantics requires an objective that jointly constrains reconstruction fidelity and semantic consistency. Thus, architectural redesigns, balanced training objectives, and specialized semantic metrics are all indispensable.

To address these, we propose Vision Foundation Model Variational Autoencoder (VFM-VAE), which leverages a frozen VFM encoder with specialized learnable modules. As shown in \cref{fig:vfmvae_arch}, our framework consists of: (1) a lightweight wrapper exposing a multi-scale hierarchy, (2) a decoder for high-fidelity reconstruction, (3) a training objective balancing semantics with pixel-level fidelity, and (4) an enhanced diagnostic metric, Semantic-Equivariant CKNNA (SE-CKNNA), which extends CKNNA to more reliably evaluate representation alignment under perturbations. We describe each component in detail below.

\subsection{VFM-VAE encoder architecture}


Unlike previous approaches that train VAE encoders from scratch to align with VFM representations, we directly leverage a pre-trained VFM as our encoder $\Phi$, \textbf{keeping it frozen throughout training} to preserve its rich semantic representation. Following insights from the literature where VFM is used for dense prediction tasks~\cite{llava, pencoder}, we recognize that optimal features for reconstruction may not reside solely in the final layer. Therefore, we extract multi-scale features from different depths of the VFM hierarchy.

Given an input image $\mathbf{x} \in \mathbb{R}^{H \times W \times 3}$, the VFM encoder extracts features in multiple layers:

\vspace{-0.5em}
\begin{equation}
\{\mathbf{f}_{\text{shallow}}, \mathbf{f}_{\text{middle}}, \mathbf{f}_{\text{final}}\} = \Phi(\mathbf{x}),
\label{eq:encode_multi_layer}
\end{equation}


\noindent where these features span the shallow, middle, and final VFM layers, capturing information from fine-grained spatial details to high-level semantics.

Similarly to the usual VAE for latent diffusion~\cite{ldm}, we need to further reduce the latent dimension to facilitate learning of the diffusion model. Hence we then concatenate these multi-scale features along the channel dimension and apply a learnable lightweight projection network, $\mathcal{C}$, which outputs the parameters of a diagonal Gaussian posterior distribution—the mean $\boldsymbol{\mu}$ and the log-variance $\log \boldsymbol{\sigma}^2$:

\vspace{-1.0em}
\begin{equation}
\boldsymbol{\mu}, \log \boldsymbol{\sigma}^2 = \mathcal{C}(\text{Concat}[\mathbf{f}_{\text{shallow}}, \mathbf{f}_{\text{middle}}, \mathbf{f}_{\text{final}}]).
\label{eq:encode_multi_concat}
\end{equation}

From this posterior, the latent code $\mathbf{z}$ is obtained via the reparameterization 
trick~\cite{vae}. We also impose a representation reconstruction loss to ensure that $\mathbf{z}$ preserves essential information from the original high-dimensional VFM representation, which will be introduced in the following subsection. This encoder design is crucial for maintaining the semantic richness of the VFM while ensuring computational efficiency for the subsequent diffusion training.

\subsection{VFM-VAE decoder architecture}

Unlike the standard SD-VAE decoder~\cite{ldm}, which maps a single latent input to a single image output, our proposed decoder architecture incorporates two key upgrades: Multi-Scale Latent Fusion and Progressive Resolution Reconstruction Blocks. These components are specifically designed to address the challenge of reconstructing high-fidelity images from semantically-rich but detail-poor VFM representations. Below, we first introduce the Multi-Scale Latent Fusion mechanism, followed by a detailed description of the Progressive Reconstruction Blocks.

\noindent \textbf{Multi-Scale latent fusion.}
\label{sec:multiscalelatent}
Given the latent representation $\mathbf{z}$, we first decompose it into global and spatial components for scale-specific processing. A global component, $\mathbf{z}_g = \text{GlobalPool}(\mathbf{z}) \in \mathbb{R}^{c}$, captures overall style information, functioning as a holistic descriptor that is invariant to spatial layout. Concurrently, a set of spatial components, $\{\mathbf{z}_s^{(i)} \in \mathbb{R}^{c \times h_i \times w_i}\}_{i=1}^{N}$, are generated by applying reshape operations to create different scales from $\mathbf{z}$. The reshape operations include techniques like pixel shuffling and unshuffling~\cite{pixelshuffle}; further implementation details are provided in the Supplementary Material Sec.~3. These decomposed latent components then serve as inputs to the subsequent progressive reconstruction blocks.

\noindent \textbf{Progressive reconstruction blocks.}
To enhance the VAE decoder’s fidelity in detail synthesis, we redesign its core architecture around progressive multi-scale reconstruction. The decoder synthesizes the image through a sequence of $N$ blocks $\{\mathcal{B}_i\}_{i=1}^N$. For a $256 \times 256$ output, these blocks operate over the resolution stages $\{8 \rightarrow 16 \rightarrow 32 \rightarrow 64 \rightarrow 128 \rightarrow 256\}$, with each 
$\mathcal{B}_i$ responsible for upsampling and refining features at its corresponding scale. The inputs to the blocks are carefully controlled, following a clear separation between global guidance from $\mathbf{z}_g$ and (in early stages) spatial injections from $\mathbf{z}_s^{(i)}$. The full block-wise formulation is provided in the Supplementary Material Sec.~3.


Structurally, the global semantic control $\mathbf{z}_g$ is obtained via global pooling across the patch dimension and supplied to \textbf{every} block. This ensures a consistent, holistic style across all scales of the synthesis process, from the initial layout to the finest textures.


In contrast to constant global control, the spatial components $\mathbf{z}_s^{(i)}$ are exclusively supplied to the \textbf{initial, low-resolution} blocks ($i \le 4$). Conceptually, this establishes the foundational layout, freeing subsequent high-resolution blocks ($5 \le i \le 6$) to focus on intricate details. Empirically, early experiments showed that higher-resolution injection via pixel shuffle yields insufficient effective channels, increasing computation without performance gains.

The core building block of our decoder is a \textbf{Modulated ConvNeXt block}, which we adapt from the standard ConvNeXt~\cite{convnext} architecture to enable fine-grained stylistic control. The global style information, provided by the global component $\mathbf{z}_g$, is injected into each block via \textbf{modulated convolutions}~\cite{stylegan, stylegan-t}. This modulation is applied at the first $1 \times 1$ point-wise convolution—the layer responsible for channel mixing. Here, the global component $\mathbf{z}_g$ is passed through a block-specific, learned affine transformation $\gamma_i$ to produce per-channel scaling factors. These factors then perform channel-wise weighting on the convolution's weights, effectively modulating the feature maps based on the global style. For a given block input $\mathbf{h}_{\text{in}}$, the complete operation, including the residual connection, is:

\vspace{-1.0em}
\begin{equation}
\mathcal{B}_i(\mathbf{h}_{\text{in}}, \mathbf{z}_g) = \text{ModConv}(\mathbf{h}_{\text{in}}, \gamma_i(\mathbf{z}_g)) + \mathbf{h}_{\text{in}}.
\label{eq:decode_block}
\end{equation}

To ensure that each block in this progressive hierarchy learns meaningful and scale-appropriate features, we introduce a direct supervision mechanism at every resolution. This is achieved by attaching a lightweight $\text{ToRGB}_i$ head to the output of each block. A critical aspect of this design is a \textbf{feature-space residual connection}, which ensures that the supervision at a finer scale is coherently informed by the structure established at coarser ones. This principle is implemented via our $\text{ToRGB}$ heads, itself a lightweight, modulated $1 \times 1$ convolution. Its operation is defined as follows, correctly handling the initial state where $\mathbf{h}^{(0)}$ is absent:

\vspace{-1.0em}
\begin{equation}
\hat{\mathbf{x}}_i =
\begin{cases}
\text{ToRGB}_i(\mathbf{h}^{(1)}, \mathbf{z}_g) & \text{if } i=1, \\
\text{ToRGB}_i(\mathbf{h}^{(i)} + \text{Upsample}(\mathbf{h}^{(i-1)}), \mathbf{z}_g) & \text{if } i > 1.
\end{cases}
\label{eq:multiscale}
\end{equation}

By forcing each block to generate a image $\hat{\mathbf{x}}_i$ via its $\text{ToRGB}_i$ head, our design enforces a learning signal at each stage of the synthesis. The feature-fusion step provides the $\text{ToRGB}_i$ head with a multi-layer context, allowing it to generate a projection based on both newly refined details from $\mathbf{h}^{(i)}$ and stable, structural information from $\mathbf{h}^{(i-1)}$.

\subsection{Training objectives}
The training objective for our VFM-VAE is designed to address a dual challenge: achieving high-fidelity image reconstruction while ensuring the compact latent $\mathbf{z}$ preserves the rich semantic information from the frozen VFM encoder. To this end, our comprehensive loss function combines several components targeting representation preservation, reconstruction fidelity, and perceptual realism: $\mathcal{L}_{\text{total}} = \lambda_{\text{rep}}\mathcal{L}_{\text{rep}} + \sum_{i=1}^{N} \lambda_i \mathcal{L}_{\text{recon}}^{(i)} + \lambda_{\text{GAN}}\mathcal{L}_{\text{GAN}} + \lambda_{\text{LPIPS}}\mathcal{L}_{\text{LPIPS}}$.

\noindent \textbf{Representation regularization loss.} 
To ensure our projected latent $\mathbf{z}$ maintains semantic alignment with the original VFM features, we employ a dual-component regularization loss, $\mathcal{L}_{\text{rep}}$. First, we include the standard Kullback-Leibler (KL) divergence loss, $\mathcal{L}_{\text{KL}}(\mathbf{z})$, a cornerstone of VAEs that regularizes the latent space distribution. Second, to explicitly enforce semantic similarity, we leverage the VF loss~\cite{vavae}, denoted as $\mathcal{L}_{\text{VF}}(\mathbf{z}, \mathbf{f}_{\text{final}})$. This loss, which computes cosine similarity and matrix distance, is specifically crafted to regularize high-dimensional representations without overly constraining their capacity. Our combined representation loss is therefore $\mathcal{L}_{\text{rep}}= \mathcal{L}_{\text{KL}} + \mathcal{L}_{\text{VF}}$.

\noindent \textbf{Multi-resolution reconstruction loss.} To ensure stable training and to impose a strong learning signal on each decoder block individually, we apply direct pixel-level supervision across all resolutions. The reconstruction loss for the $i$-th block is an L1 loss: $\mathcal{L}^{(i)}_{\text{recon}} = \lVert \mathrm{f}_{r_i}(\mathbf{x})- \hat{\mathbf{x}}_i \rVert_1$, where $\mathrm{f}_{r_i}$ is an operation that downsamples the ground-truth image $\mathbf{x}$ to match the resolution of the block's output, $\hat{\mathbf{x}}_i$. This multi-scale supervision is crucial for preventing mode collapse in early stage and ensuring each block learns its role in the hierarchical synthesis.

\noindent \textbf{Adversarial and perceptual losses.} To transcend the limitations of pixel-wise metrics and enhance the perceptual quality and realism of the final, full-resolution output, we incorporate two complementary losses. First, inspired by recent advances in generative modeling~\cite{stylegan-t,sdxl-turbo}, we employ an adversarial loss $\mathcal{L}_{\text{GAN}}$. This involves a discriminator with a DINOv2-based backbone~\cite{dinov2}. Second, we include the LPIPS perceptual loss~\cite{lpips}, $\mathcal{L}_{\text{LPIPS}}$, which more closely aligns with human perception of image similarity.

Overall, the proposed VFM-VAE effectively balances semantic quality with reconstruction fidelity, {offering a strong foundation for downstream diffusion training.}

\subsection{Semantic-Equivariant CKNNA}
\label{sec:se_cknna}




REPA~\cite{repa} identifies the development of internal feature structures as a primary challenge in diffusion training. To quantify this process, REPA leverages CKNNA~\cite{platonic} to measure representation quality via VFM-alignment. This framework establishes that better VFM-alignment yields superior generation, an insight widely adopted by subsequent works such as REPA-E~\cite{repae} and REG~\cite{reg}. Building on this, we apply CKNNA to investigate tokenizer-side alignment and its impact on the internal feature evolution of diffusion models. Technically, CKNNA is proposed to measure the preservation of local neighborhood structures between two representation spaces.

Our empirical analysis (\cref{fig:brittleness}) reveals that existing distillation-based tokenizers exhibit substantial CKNNA degradation under perturbations, despite achieving high scores on clean images. This suggests that standard CKNNA may fail to accurately capture the brittleness of aligned representations. To address this, we introduce Semantic-Equivariant CKNNA (SE-CKNNA) as a more rigorous theoretical tool to evaluate representation robustness under perturbations. By averaging alignment scores across a distribution of semantic-preserving transformations, SE-CKNNA provides a more reliable measure of a tokenizer's alignment stability.

\begin{table*}[t]
\centering
\caption{Reconstruction, generation, and representation metrics. The generation results are obtained by training LightningDiT-XL~\cite{vavae} on ImageNet. \#Images indicates the number of training images used for each tokenizer. Top-1 Acc. denotes the result of linear probing. Details of SE-CKNNA are provided in \cref{sec:se_cknna}. CKNNA and SE-CKNNA are computed with their respective aligned VFMs (DINOv2-Large for SD-VAE). \textit{Relative Change} represents the variation between SE-CKNNA and CKNNA, calculated as $|\text{SE-CKNNA} - \text{CKNNA}| / \text{CKNNA}$.}
\label{tab:vae_metrics}
\resizebox{\linewidth}{!}{%
\begin{tabular}{@{}c|c|cc|ccc|cccc@{}}
\toprule
\multirow{2}{*}{Tokenizer} & \multirow{2}{*}{\#Images} & \multicolumn{2}{c|}{\textit{Reconstruction}} & \multicolumn{3}{c|}{\textit{Generation}} & \multicolumn{4}{c}{\textit{Representation Metrics}} \\
 &  & rFID$\downarrow$ & rIS$\uparrow$ & Epochs & gFID$\downarrow$  & gIS$\uparrow$ & Top-1 Acc.$\uparrow$ & CKNNA & SE-CKNNA & Relative Change \\ \midrule
 SD-VAE & 108M & 0.62 & 212.1 & 80 & 7.13 & - & 8.0 & 0.004 & 0.005 & - \\
 VA-VAE & 160M & \textbf{0.30} & 213.6 & 64 & 5.14 & 130.2 & 31.9 & 0.202 & 0.135 & $-33.2\%$ \\
 VFM-VAE & 44M & 0.52 & \textbf{214.1} & 64 & \textbf{3.80} & \textbf{152.8} & \textbf{43.2} & 0.188 & 0.191 & $+1.6\%$ \\ \bottomrule
\end{tabular}%
}
\end{table*}







\noindent \textbf{Definition.} SE-CKNNA serves as a Monte Carlo-style estimate of alignment over a semantic-preserving perturbation distribution $\mathcal{T}$, simply obtained by averaging CKNNA scores across sampled transformations:

\begin{equation}
\mathrm{SE\mbox{-}CKNNA} = \frac{1}{|\mathcal{T}|} \sum_{T\in\mathcal{T}} \mathrm{CKNNA}(T).\label{eq:se_cknna_main}
\end{equation}

In practice, we uniformly sample from a set of representative equivariant transformations that modify spatial arrangement or pixel values without altering core semantics:

\begin{itemize}[leftmargin=*,itemsep=0mm]
\item Additive noise: Strengths $\{0.05, 0.10, 0.15, 0.20\}$ (in the $[0,1]$ range).
\item Scale interpolations: Ratios $\{0.25, 0.50, 0.75, 1.0\}$.
\item Discrete rotations: $\{0^\circ, 90^\circ, 180^\circ, 270^\circ\}$.
\end{itemize}

Visualization of the transformed samples and further formulations are provided in Supplementary Material Sec.~2.

%% file: sec_camera_ready/4_experiments_compact_for_arxiv.tex
\begin{figure}[t]
\centering
\includegraphics[width=\linewidth]{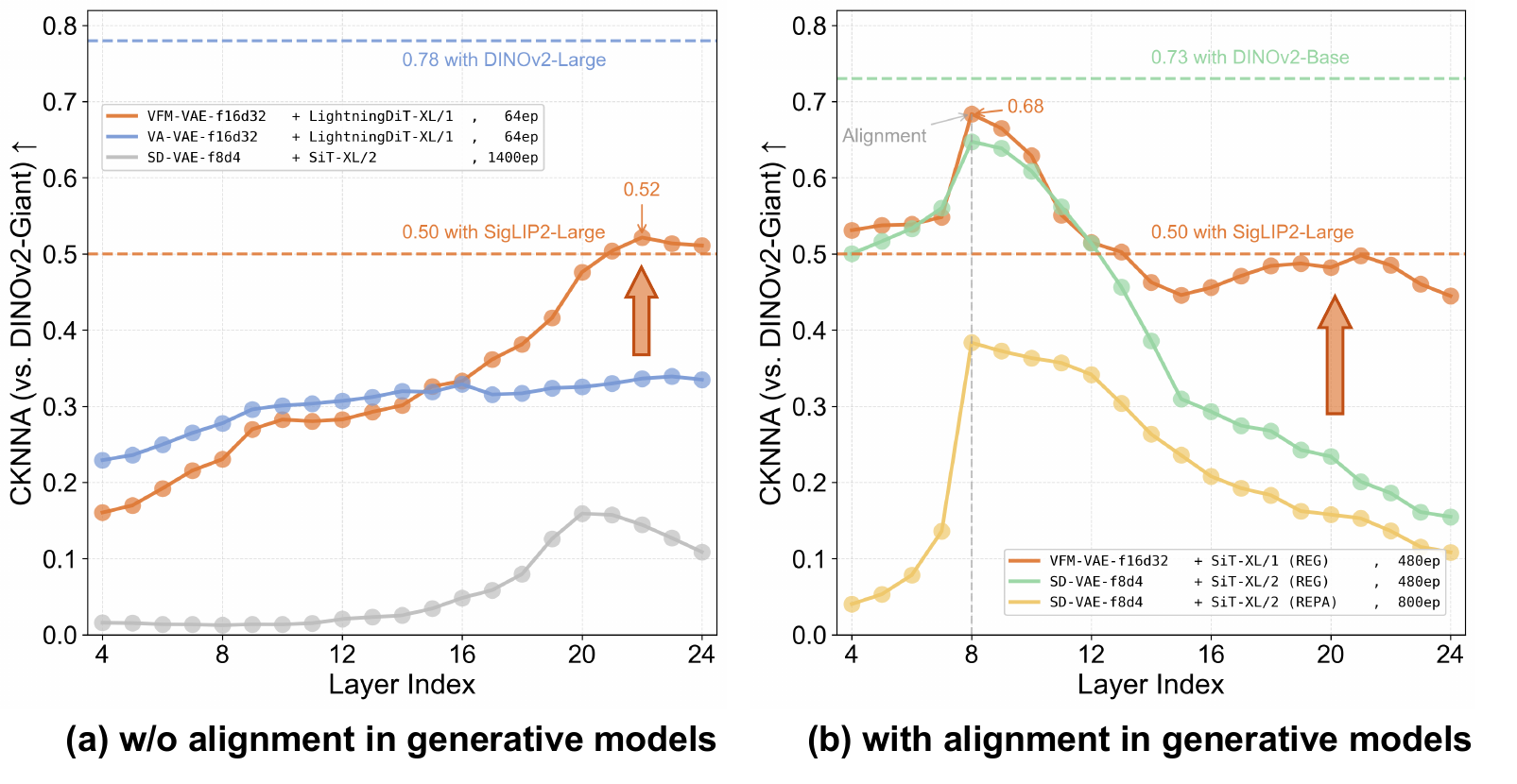}
\caption{CKNNA comparison across layers of the diffusion models.
(a) Without explicit VFM alignment, the diffusion model combined with VFM-VAE achieves higher average and peak layer-wise CKNNA than other tokenizer baselines.
(b) With alignment enabled, the VFM-VAE system further improves, consistently surpassing other diffusion-aligned baselines in CKNNA.}
\label{fig:layerwise-compact}
\end{figure}

\section{Experiments}
\label{sec:experiments}


In this section, we first introduce our experiment setup. Then we examine VFM-VAE as an independent tokenizer, assessing its reconstruction fidelity and alignment with VFM representations. Next, we systematically analyze how different tokenizer representations impact representation learning within diffusion models. Finally, we demonstrate how integrating VFM-VAE into diffusion pipelines leads to consistently stronger generative performance.

\input{tables/main_comparison_compact}

\subsection{Setup}
\label{sec:setup}

\noindent \textbf{Baselines.} Our analysis covers some representative models. {For visual tokenizers}, we benchmark VFM-VAE against SD-VAE~\cite{ldm} and alignment-based VAEs, including VA-VAE~\cite{vavae} and REPA-E~\cite{repae}. {For generative models}, we compare with REPA~\cite{repa} and REG~\cite{reg}, both designed to align diffusion transformer features with those of a VFM. Together, these tokenizers and generative models form the most relevant and competitive baselines for evaluation.

\noindent \textbf{Implementation details.}
All tokenizer and generative model training is conducted on ImageNet~\cite{imagenet} at 256×256 resolution. VFM-VAE adopts the same f16d32 configuration as VA-VAE for fair comparison. VA-VAE is aligned with DINOv2-Large~\cite{dinov2}, while VFM-VAE uses SigLIP2-Large~\cite{siglip2} by default (see the Supplementary Material Sec.~4 for rationale). For generative models, we evaluate two settings: (1) LightningDiT-XL, matching the VA-VAE baseline, and (2) REG, which employs a SiT-XL backbone~\cite{sit} with an additional alignment loss. Full training details are provided in the 
Supplementary Material Sec.~5.

\noindent \textbf{Evaluation.} For reconstruction performance, we report Fréchet Inception Distance (FID)~\cite{fid} and Inception Score (IS)~\cite{is} on the full ImageNet 50K validation set. For generative performance, we follow the ADM~\cite{adm} setup and report generation FID (gFID), spatial FID (sFID)~\cite{sfid}, IS, as well as precision (Prec.) and recall (Rec.)~\cite{precision}. For alignment analysis, we follow the original protocol~\cite{platonic} of CKNNA to calculate on representation.

\subsection{Reconstruction and Alignment Evaluation}

\noindent \textbf{VFM-VAE achieves semantic-rich representations and strong reconstruction fidelity.}
Reconstruction quality sets the ceiling for downstream diffusion generation. VFM-VAE performs strongly on this front: it preserves object structure and natural textures, and on quantitative metrics (\cref{tab:vae_metrics}) its semantic–spatial dual-branch design \textbf{surpasses SD-VAE} on fidelity measures such as FID and IS. Additional ablations appear in \cref{sec:ablation}, with qualitative reconstructions in the Supplementary Material Sec.~6.


Beyond reconstruction, we also evaluate representation alignment with CKNNA and our proposed SE-CKNNA~(\cref{sec:se_cknna}). As shown in \cref{fig:brittleness}, VFM-VAE exhibits more stable alignment: with only \textbf{25\% of the training images} used by VA-VAE, its SE-CKNNA differs from CKNNA by just +1.6\%, in contrast to VA-VAE’s –33.2\% drop. Additionally, we validated the linear probing results, where VFM-VAE significantly outperforms VA-VAE, improving from 31.9\% to 43.2\%. These results highlight that inheriting frozen VFM features yields a more robust and generalizable latent space that maintains semantics under perturbations. This finding raises a question: \emph{Could a tokenizer with more stable alignment to the VFM enable stronger representation learning within diffusion models?} We examine this in the next paragraph.

\subsection{Impact Analysis on Diffusion Models}
\label{sec:impact-diffusion}

\noindent \textbf{High-quality tokenizer representation facilitates diffusion representation learning.}
Prior work such as REPA~\cite{repa}, REPA-E~\cite{repae}, and REG~\cite{reg} shows that layer-wise alignment with a VFM is closely tied to representation learning in diffusion models. Following this perspective, in \cref{fig:layerwise-compact} we compare the layer-wise CKNNA of diffusion models based on VFM-VAE, VA-VAE and SD-VAE, both trained without explicit alignment. The diffusion model based on VFM-VAE achieves higher average and peak CKNNA, surpassing its VA-VAE–based counterpart as training progresses. Its peak value reaches 0.52, above the 0.50 reference computed between SigLIP2-Large and DINOv2-Giant, indicating more effective inheritance of VFM feature structure. In terms of generation quality (~\cref{tab:vae_metrics}), LightningDiT trained with VFM-VAE attains a 3.80 FID at 64 epochs without classifier-free guidance (CFG)~\cite{cfg}, outperforming the VA-VAE–based model’s 5.14. These results answer the question raised at the end of the previous paragraph, showing diffusion models based on VFM-VAE develop stronger internal representations.


\noindent \textbf{Synergistic benefits of dual-side alignment on tokenizer and diffusion.}
Building on strong results without explicit alignment, we further explore the synergy between VFM-aligned tokenizers and diffusion models. As shown in \cref{fig:layerwise-compact}, when paired with LightningDiT, shallow-layer alignment remains weak, with limited early-layer CKNNA improvements; by the 16th layer, however, the VFM-VAE–based model surpasses the VA-VAE–based baseline in alignment quality. Ideally, a diffusion model should maintain high-quality representation throughout, raising the question: \textit{can shallow layers be explicitly aligned to ensure consistency across all depths?}

To explore this, we incorporate REG~\cite{reg}, which aligns shallow layers with VFM patch features and injects global semantics via the VFM class token. As shown in \cref{fig:layerwise-compact}, combining tokenizer-side alignment (VFM-VAE) with shallow-layer alignment (REG) produces consistently strong layer-wise CKNNA across all depths, yielding higher and more uniform alignment than REG or REPA alone. This alignment advantage finally derives improved generative performance. We discuss it in the next subsection.

\begin{table}[t]
\centering
\caption{Comparison of ImageNet-512 generation performance after 100k training steps without CFG using LightningDiT-B/1. VFM-VAE demonstrates improved generation quality over the VA-VAE baseline. Prec. and Rec. denote precision and recall.}
\label{tab:higher-resolution-compact}
\vspace{-0.5em}
\resizebox{\linewidth}{!}{%
\begin{tabular}{@{}l|ccccc@{}}
\toprule
Method                     & gFID$\downarrow$ & gIS$\uparrow$ & Prec.$\uparrow$ & Rec.$\uparrow$ \\ \midrule
VA-VAE + LightningDiT-B/1  & 21.42   & 55.3   & 0.75     & \textbf{0.60}    \\
VFM-VAE + LightningDiT-B/1 & \textbf{18.05} & \textbf{69.6}   & \textbf{0.78}     & \textbf{0.60}    \\ \bottomrule
\end{tabular}%
}
\vspace{-1em}
\end{table}

\begin{table}[t]
\centering
\caption{Text-to-image generation results on DPG-Bench~\cite{dpgbench} and MJHQ-30K~\cite{mjhq30k} using BLIP3-o~\cite{blip3o} (256px, 1 pretraining epoch). VFM-VAE + BLIP3-o outperforms the VA-VAE variant.}
\label{tab:ablation-t2i-compact}
\vspace{-0.5em}
\resizebox{\linewidth}{!}{
\begin{tabular}{@{}l|cc@{}}
\toprule
\multirow{2}{*}{Text-to-image Model} & \multicolumn{2}{c}{\textit{Overall metrics}}   \\
                                     & score (DPG-Bench)$\uparrow$ & gFID (MJHQ-30K)$\downarrow$ \\ \midrule
VA-VAE + BLIP3-o    & 55.4         & 23.0        \\
VFM-VAE + BLIP3-o   & \textbf{59.1}  & \textbf{17.0}                 \\ \bottomrule
\end{tabular}
}
\vspace{-1em}
\end{table}

\subsection{Generation performance}


Tab.~\cref{tab:main} presents the systematic comparison. VFM-VAE + REG reaches 2.22 gFID (w/o CFG~\cite{cfg}) in 64 epochs, matching 480-epoch REG and highlighting the synergy of dual-side alignment. \textbf{At 640 epochs, our model attains 1.62/1.31 FIDs (w/o and w/ CFG), demonstrating faster convergence and superior quality.} Even without diffusion alignment, VFM-VAE + LightningDiT outperforms the VA-VAE variant by 1.34 gFID. Further ablations on ImageNet-512 and text-to-image (\cref{tab:higher-resolution-compact,tab:ablation-t2i-compact}) confirm VFM-VAE’s performance lead over VA-VAE. Detailed implementation and expanded numerical results for these two ablation studies are included in Supp. Sec.~4, while stage-wise qualitative comparisons are presented in Sec.~6.

\begin{table}[t]
\caption{VFM-VAE module ablation studies. Modules are added while maintaining 5M-image weak alignment to VFM.}
\label{tab:ablation_components}
\vspace{-0.5em}
\centering
\resizebox{0.75\linewidth}{!}{
\begin{tabular}{lcc}
\toprule
Setting & rFID$\downarrow$ & rIS$\uparrow$ \\
\midrule
SD-VAE-style Baseline & 19.69 & 74.9 \\
+ Multi-scale Latent Fusion & 14.35 & 93.6 \\
+ Our Modern Blocks & 1.08 & 194.6 \\
+ Encoder Modifications & \textbf{0.71} & \textbf{206.8} \\
\bottomrule
\end{tabular}}
\vspace{-1.5em}
\end{table}

\begin{table}[t]
\centering
\caption{Comparison of VFM-VAE variants across different VFMs. Each tokenizer is trained through two-stage alignment (5M strong/weak). Generation results are reported at 100k training steps using LightningDiT-L/1 without CFG.}
\label{tab:ablation-vfm-compact}
\vspace{-0.5em}
\resizebox{\linewidth}{!}{%
\begin{tabular}{@{}l|ccc|cc@{}}
\toprule
\multirow{2}{*}{VFM} & \multicolumn{3}{c|}{\textit{Reconstruction}}                                               & \multicolumn{2}{c}{\textit{Generation}}                                                 \\
                     & rFID$\downarrow$          & rIS$\uparrow$             & PSNR$\uparrow$            & gFID$\downarrow$           & gIS$\uparrow$             \\ \midrule
EVA-CLIP-Large~\cite{eva-clip}       & \textbf{1.35} & 188.4          & \textbf{19.33}& 4.40 & 146.4 \\
DINOv2-Large~\cite{dinov2}         & 1.55          & \textbf{199.8} & 17.60 & \textbf{4.00} & \textbf{147.1} \\
SigLIP2-Large~\cite{siglip2}        & 1.61          & 178.0          & 18.73 & 5.59 & 127.8 \\ \bottomrule
\end{tabular}%
}
\vspace{-1.5em}
\end{table}

\subsection{Ablation studies}
\label{sec:ablation}

The VFM-VAE consists of a frozen VFM, a lightweight encoder, and a decoder, using a dual-branch design for semantic/spatial control and a set of losses to maintain VFM alignment. We begin from a minimal baseline and progressively add components to assess their impact on reconstruction:

\begin{itemize}[leftmargin=*,itemsep=0mm]
    \item \textbf{SD-VAE-style Baseline.} We start from an SD-VAE-style design using SigLIP2-Large-Patch16-256~\cite{siglip2} as the VFM. The encoder and decoder each use two convolutional layers for pre- and post-sampling, with L1, LPIPS, KL, and adversarial losses.
    
    \item \textbf{Multi-scale Latent Fusion.} We add Multi-scale Latent Fusion (\cref{sec:multiscalelatent}), upgrade the upsampling modules, and introduce multi-scale reconstruction losses to stabilize training and speed up early convergence.
    
    \item \textbf{Our Modern Blocks.} We replace modulated convolutions with Modulated ConvNeXt Block and add self-attention at low resolutions for improved decoding.
    
    \item \textbf{Encoder Modifications.} We aggregate shallow, intermediate, and final VFM features to capture multi-level semantics and upgrade the backbone to SigLIP2-Large-Patch16-512~\cite{siglip2} for finer detail representation.
\end{itemize}


As shown in \cref{tab:ablation_components}, a minimal baseline trained on 5M images yields an rFID of 19.79. Adding spatial control reduces this by~27\%, while enhanced capacity and modern modules further bring rFID to~1.08, with final refinements pushing it below~1.0. These results demonstrate that each component improves reconstruction quality while preserving VFM alignment. Conversely, scaling up an SD-VAE-style baseline (\cref{tab:vae_metrics}) leads to decreased performance. \cref{tab:ablation-vfm-compact} demonstrates VFM-VAE’s broad compatibility, yielding robust reconstruction and generation performance across different VFMs. Detailed results for these ablations and VFM compatibility studies are in Supp. Sec.~4.

%% file: tables/main_comparison_compact.tex
\begin{table*}[t]
\centering
\caption{System-level generative performance on ImageNet 256×256. We evaluated VFM-VAE against leading VAEs used as tokenizers for LDM training. Results show that VFM-VAE offers a stronger latent foundation, enabling models to consistently reach higher performance. A standout case is its pairing with REG~\cite{reg}, achieving a gFID of 1.62 in 640 epochs without classifier-free guidance~\cite{cfg}.}
\label{tab:main}
\resizebox{\textwidth}{!}{%
\begin{tabular}{@{}cccccccccccc@{}}
\toprule
\multicolumn{1}{c|}{} & \multicolumn{1}{c|}{} & \multicolumn{1}{c|}{} & \multicolumn{1}{c|}{} & \multicolumn{1}{c|}{} & \multicolumn{2}{c|}{Generation w/o CFG} & \multicolumn{5}{c}{Generation w/ CFG} \\ \cmidrule(l){6-12} 
\multicolumn{1}{c|}{\multirow{-2}{*}{Tokenizer}} & \multicolumn{1}{c|}{\multirow{-2}{*}{Generative Model}} & \multicolumn{1}{c|}{\multirow{-2}{*}{Training Epoches}} & \multicolumn{1}{c|}{\multirow{-2}{*}{\#params}} & \multicolumn{1}{c|}{\multirow{-2}{*}{rFID$\downarrow$}} & gFID$\downarrow$ & \multicolumn{1}{c|}{gIS$\uparrow$} & gFID$\downarrow$ & sFID$\downarrow$ & gIS$\uparrow$ & Prec.$\uparrow$ & Rec.$\uparrow$ \\ \midrule
\multicolumn{12}{c}{\cellcolor[HTML]{EBEBFF}\textbf{AutoRegressive (AR)}} \\ \midrule
\multicolumn{1}{c|}{MaskGiT} & \multicolumn{1}{c|}{MaskGiT~\cite{maskgit}} & \multicolumn{1}{c|}{555} & \multicolumn{1}{c|}{227M} & \multicolumn{1}{c|}{2.28} & 6.18 & \multicolumn{1}{c|}{182.1} & - & - & - & - & - \\
\multicolumn{1}{c|}{VQGAN~\cite{vqgan}} & \multicolumn{1}{c|}{LlamaGen~\cite{llamagen}} & \multicolumn{1}{c|}{300} & \multicolumn{1}{c|}{3.1B} & \multicolumn{1}{c|}{0.59} & 9.38 & \multicolumn{1}{c|}{112.9} & 2.18 & 5.97 & 263.3 & 0.81 & 0.58 \\
\multicolumn{1}{c|}{VQVAE~\cite{vavae}} & \multicolumn{1}{c|}{VAR~\cite{var}} & \multicolumn{1}{c|}{350} & \multicolumn{1}{c|}{2.0B} & \multicolumn{1}{c|}{-} & - & \multicolumn{1}{c|}{-} & 1.80 & - & \textbf{365.4} & \textbf{0.83} & 0.57 \\
\multicolumn{1}{c|}{LFQ tokenizers} & \multicolumn{1}{c|}{MagViT-v2~\cite{magvitv2}} & \multicolumn{1}{c|}{1080} & \multicolumn{1}{c|}{307M} & \multicolumn{1}{c|}{1.50} & 3.65 & \multicolumn{1}{c|}{200.5} & 1.78 & - & 319.4 & - & - \\
\multicolumn{1}{c|}{LDM~\cite{ldm}} & \multicolumn{1}{c|}{MAR~\cite{mar}} & \multicolumn{1}{c|}{800} & \multicolumn{1}{c|}{945M} & \multicolumn{1}{c|}{0.53} & 2.35 & \multicolumn{1}{c|}{227.8} & 1.55 & - & 303.7 & 0.81 & 0.62 \\ \midrule
\multicolumn{12}{c}{\cellcolor[HTML]{EBEBFF}\textbf{Latent Diffusion Models (LDM)}} \\ \midrule
\multicolumn{1}{c|}{} & \multicolumn{1}{c|}{MaskDiT~\cite{maskdit}} & \multicolumn{1}{c|}{1600} & \multicolumn{1}{c|}{675M} & \multicolumn{1}{c|}{} & 5.69 & \multicolumn{1}{c|}{177.9} & 2.28 & 5.67 & 276.6 & 0.80 & 0.61 \\
\multicolumn{1}{c|}{} & \multicolumn{1}{c|}{DiT~\cite{dit}} & \multicolumn{1}{c|}{1400} & \multicolumn{1}{c|}{675M} & \multicolumn{1}{c|}{} & 9.62 & \multicolumn{1}{c|}{121.5} & 2.27 & 4.60 & 278.2 & \textbf{0.83} & 0.57 \\
\multicolumn{1}{c|}{} & \multicolumn{1}{c|}{SiT~\cite{sit}} & \multicolumn{1}{c|}{1400} & \multicolumn{1}{c|}{675M} & \multicolumn{1}{c|}{} & 8.61 & \multicolumn{1}{c|}{131.7} & 2.06 & 4.50 & 270.3 & 0.82 & 0.59 \\
\multicolumn{1}{c|}{} & \multicolumn{1}{c|}{FasterDiT~\cite{fasterdit}} & \multicolumn{1}{c|}{400} & \multicolumn{1}{c|}{675M} & \multicolumn{1}{c|}{} & 7.91 & \multicolumn{1}{c|}{131.3} & 2.03 & 4.63 & 264.0 & 0.81 & 0.60 \\
\multicolumn{1}{c|}{} & \multicolumn{1}{c|}{MDT~\cite{mdt}} & \multicolumn{1}{c|}{1300} & \multicolumn{1}{c|}{675M} & \multicolumn{1}{c|}{} & 6.23 & \multicolumn{1}{c|}{143.0} & 1.79 & 4.57 & 283.0 & 0.81 & 0.61 \\
\multicolumn{1}{c|}{\multirow{-6}{*}{SD-VAE~\cite{ldm}}} & \multicolumn{1}{c|}{MDTv2~\cite{mdtv2}} & \multicolumn{1}{c|}{1080} & \multicolumn{1}{c|}{675M} & \multicolumn{1}{c|}{\multirow{-6}{*}{0.62}} & - & \multicolumn{1}{c|}{-} & 1.58 & 4.52 & 314.7 & 0.79 & 0.65 \\ \midrule
\multicolumn{12}{c}{\cellcolor[HTML]{EBEBFF}\textbf{Representation Alignment Methods}} \\ \midrule
\multicolumn{1}{c|}{} & \multicolumn{1}{c|}{} & \multicolumn{1}{c|}{80} & \multicolumn{1}{c|}{675M} & \multicolumn{1}{c|}{} & 3.46 & \multicolumn{1}{c|}{159.8} & 1.67 & 4.12 & 266.3 & 0.80 & 0.63 \\
\multicolumn{1}{c|}{\multirow{-2}{*}{E2E-VAE~\cite{repae}}} & \multicolumn{1}{c|}{\multirow{-2}{*}{REPA~\cite{repa}}} & \multicolumn{1}{c|}{800} & \multicolumn{1}{c|}{675M} & \multicolumn{1}{c|}{\multirow{-2}{*}{\textbf{0.28}}} & 1.83 & \multicolumn{1}{c|}{217.3} & \textbf{1.26} & \textbf{4.11} & 314.9 & 0.79 & \textbf{0.66} \\ \midrule
\multicolumn{1}{c|}{} & \multicolumn{1}{c|}{} & \multicolumn{1}{c|}{64} & \multicolumn{1}{c|}{675M} & \multicolumn{1}{c|}{} & 5.14 & \multicolumn{1}{c|}{130.2} & 2.11 & 4.16 & 252.3 & 0.81 & 0.58 \\
\multicolumn{1}{c|}{} & \multicolumn{1}{c|}{} & \multicolumn{1}{c|}{80} & \multicolumn{1}{c|}{675M} & \multicolumn{1}{c|}{} & 4.29 & \multicolumn{1}{c|}{-} & - & - & - & - & - \\
\multicolumn{1}{c|}{\multirow{-3}{*}{VA-VAE~\cite{vavae}}} & \multicolumn{1}{c|}{\multirow{-3}{*}{LightningDiT~\cite{vavae}}} & \multicolumn{1}{c|}{800} & \multicolumn{1}{c|}{675M} & \multicolumn{1}{c|}{\multirow{-3}{*}{0.30}} & 2.17 & \multicolumn{1}{c|}{205.6} & 1.35 & 4.15 & 295.3 & 0.79 & 0.65 \\ \midrule
\multicolumn{1}{c|}{} & \multicolumn{1}{c|}{} & \multicolumn{1}{c|}{80} & \multicolumn{1}{c|}{675M} & \multicolumn{1}{c|}{} & 7.90 & \multicolumn{1}{c|}{122.6} & - & - & - & - & - \\
\multicolumn{1}{c|}{} & \multicolumn{1}{c|}{\multirow{-2}{*}{REPA}} & \multicolumn{1}{c|}{800} & \multicolumn{1}{c|}{675M} & \multicolumn{1}{c|}{} & 5.90 & \multicolumn{1}{c|}{157.8} & 1.42 & 4.70 & 305.7 & 0.80 & 0.65 \\ \cmidrule(lr){2-4} \cmidrule(l){6-12} 
\multicolumn{1}{c|}{} & \multicolumn{1}{c|}{} & \multicolumn{1}{c|}{80} & \multicolumn{1}{c|}{675M} & \multicolumn{1}{c|}{} & 3.40 & \multicolumn{1}{c|}{184.1} & 1.86 & 4.49 & 321.4 & 0.76 & 0.63 \\
\multicolumn{1}{c|}{\multirow{-4}{*}{SD-VAE}} & \multicolumn{1}{c|}{\multirow{-2}{*}{REG~\cite{reg}}} & \multicolumn{1}{c|}{480} & \multicolumn{1}{c|}{675M} & \multicolumn{1}{c|}{\multirow{-4}{*}{0.62}} & 2.20 & \multicolumn{1}{c|}{219.1} & 1.40 & 4.24 & 296.9 & 0.77 & \textbf{0.66} \\ \midrule
\multicolumn{1}{c|}{} & \multicolumn{1}{c|}{} & \multicolumn{1}{c|}{64} & \multicolumn{1}{c|}{675M} & \multicolumn{1}{c|}{} & \cellcolor{lightyellow}3.80 & \multicolumn{1}{c|}{\cellcolor{lightyellow}152.8} & \cellcolor{lightyellow}2.16 & \cellcolor{lightyellow}4.26 & \cellcolor{lightyellow}232.8 & \cellcolor{lightyellow}0.82 & \cellcolor{lightyellow}0.58 \\
\multicolumn{1}{c|}{} & \multicolumn{1}{c|}{} & \multicolumn{1}{c|}{80} & \multicolumn{1}{c|}{675M} & \multicolumn{1}{c|}{} & \cellcolor{lightyellow}3.41 & \multicolumn{1}{c|}{\cellcolor{lightyellow}160.4} & \cellcolor{lightyellow}- & \cellcolor{lightyellow}- & \cellcolor{lightyellow}- & \cellcolor{lightyellow}- & \cellcolor{lightyellow}- \\
\multicolumn{1}{c|}{} & \multicolumn{1}{c|}{\multirow{-3}{*}{LightningDiT}} & \multicolumn{1}{c|}{560} & \multicolumn{1}{c|}{675M} & \multicolumn{1}{c|}{} & \cellcolor{lightyellow}2.06 & \multicolumn{1}{c|}{\cellcolor{lightyellow}205.8} & \cellcolor{lightyellow}1.57 & \cellcolor{lightyellow}4.56 & \cellcolor{lightyellow}254.4 & \cellcolor{lightyellow}0.80 & \cellcolor{lightyellow}0.64 \\ 
\cmidrule(lr){2-4} \cmidrule(l){6-12} 
\multicolumn{1}{c|}{} & \multicolumn{1}{c|}{} & \multicolumn{1}{c|}{64} & \multicolumn{1}{c|}{685M} & \multicolumn{1}{c|}{} & \cellcolor{lightyellow}2.42 & \multicolumn{1}{c|}{\cellcolor{lightyellow}215.2} & \cellcolor{lightyellow}2.03 & \cellcolor{lightyellow}5.23 & \cellcolor{lightyellow}261.7 & \cellcolor{lightyellow}\textbf{0.83} & \cellcolor{lightyellow}0.58 \\
\multicolumn{1}{c|}{} & \multicolumn{1}{c|}{} & \multicolumn{1}{c|}{80} & \multicolumn{1}{c|}{685M} & \multicolumn{1}{c|}{} & \cellcolor{lightyellow}2.22 & \multicolumn{1}{c|}{\cellcolor{lightyellow}218.8} & \cellcolor{lightyellow}- & \cellcolor{lightyellow}- & \cellcolor{lightyellow}- & \cellcolor{lightyellow}- & \cellcolor{lightyellow}- \\
\multicolumn{1}{c|}{} & \multicolumn{1}{c|}{} & \multicolumn{1}{c|}{480} & \multicolumn{1}{c|}{685M} & \multicolumn{1}{c|}{} & \cellcolor{lightyellow}1.67 & \multicolumn{1}{c|}{\cellcolor{lightyellow}238.3} & \cellcolor{lightyellow}1.34 & \cellcolor{lightyellow}4.59 & \cellcolor{lightyellow}302.7 & \cellcolor{lightyellow}0.78 & \cellcolor{lightyellow}0.65 \\
\multicolumn{1}{c|}{\multirow{-7}{*}{VFM-VAE}} & \multicolumn{1}{c|}{\multirow{-4}{*}{REG}} & \multicolumn{1}{c|}{640} & \multicolumn{1}{c|}{685M} & \multicolumn{1}{c|}{\multirow{-7}{*}{0.52}} & \cellcolor{lightyellow}\textbf{1.62} & \multicolumn{1}{c|}{\cellcolor{lightyellow}\textbf{241.6}} & \cellcolor{lightyellow}1.31 & \cellcolor{lightyellow}4.63 & \cellcolor{lightyellow}300.2 & \cellcolor{lightyellow}0.78 & \cellcolor{lightyellow}\textbf{0.66} \\ \bottomrule
\end{tabular}%
}
\vspace{-1em}
\end{table*}

%% file: sec_camera_ready/5_conclusion.tex


\section{Conclusion and Limitations}
\label{sec:conclusion}

We introduced VFM-VAE, directly integrating frozen VFMs into the latent diffusion pipeline to avoid representation degradation in distillation-based alignment. We also proposed SE-CKNNA, a robust metric for evaluating alignment under perturbations. While providing strong semantic alignment and reconstruction, VFM-VAE faces limitations: leveraging frozen VFMs sacrifices some high-frequency fidelity, and inheriting complex training objectives from prior works complicates tuning. Moreover, SE-CKNNA’s scores remain relative to the specific aligned VFMs used for comparison. Overall, the empirical evidence suggests that frozen VFMs, when used with their native semantics preserved, can serve as strong and practical visual tokenizers for latent diffusion models.

%% file: sec_camera_ready/6_acknowledgments.tex
\section*{Acknowledgments}

This work was supported by the Fundamental and Interdisciplinary Disciplines Breakthrough Plan of the Ministry of Education of China under Grant JYB2025XDXM504.

%% file: sec_camera_ready/X_suppl.tex
\clearpage
\appendix
\maketitlesupplementary

\input{sec_camera_ready/X_suppl_1_relation}
\input{sec_camera_ready/X_suppl_2_cknna}
\input{sec_camera_ready/X_suppl_3_arch}
\input{sec_camera_ready/X_suppl_4_analysis}
\input{sec_camera_ready/X_suppl_5_implementation}
\input{sec_camera_ready/X_suppl_6_qualitation}

%% file: sec_camera_ready/X_suppl_1_relation.tex

\noindent \textbf{Overview.}
This supplementary material provides additional technical details, analyses, and results that complement the main paper.
\begin{itemize}
    \item \cref{appendix:relation-rae} discuss the relation with concurrent works, represented by RAE~\cite{rae} and SVG~\cite{svg}, clarifying the different design choice.
    \item \cref{appendix:alignment-metrics} summarizes the CKNNA and SE-CKNNA metrics and provides the detailed evaluation setup used throughout our alignment analyses.
    \item \cref{appendix:arch} describes the architecture of VFM-VAE, including the latent projection, global or spatial branches, and detailed progressive decoder design.
    \item \cref{appendix:analysis} presents additional experimental results: including detailed tokenizer reconstruction, representation alignment, and ablations; VFM-VAE compatibility across various VFMs; comparison with the fair VA-VAE baseline; and preliminary performance in high-resolution and unified text-to-image generation.
    \item \cref{appendix:details} lists key architectural configurations, training hyperparameters for all stages, and training efficiency.
    \item \cref{appendix:qualitative} provides extended qualitative visualizations, including tokenizer reconstructions, stage-wise generation visualizations, and sample generation results.
\end{itemize}

\section{The relation with concurrent works}
\label{appendix:relation-rae}



We note that concurrent works, such as RAE~\cite{rae} and SVG~\cite{svg}, share our core insight of leveraging a frozen Vision Foundation Model (VFM) as a tokenizer for LDMs. However, while RAE and SVG directly utilize high-dimensional VFM features, we strictly adhere to the latent channel compression characteristic of standard LDMs. Specifically, RAE focuses on enhancing the diffusion process through latent constraints, modified schedules, and autoguidance. In contrast, we shift the focus to specialized decoder architecture and loss design to bridge the gap between semantic-rich VFM features and pixel-accurate synthesis. Our VFM-VAE design offers the following advantages:

\begin{itemize}
\item \textbf{Systematic analysis of VFM-enhanced LDM training.}  We provide empirical insights into how VFM-based encoders improve representation quality across all LDM layers (Fig.~4 of our main body). Our analysis establishes that dual-side alignment, applied simultaneously to tokenizers and diffusion models, produces superior VFM-aligned representations within the diffusion framework.
\item \textbf{Seamless integration with existing VAE-LDM infrastructure.} By maintaining standard latent dimensions and projection layers, VFM-VAE remains fully compatible with the established VAE-LDM ecosystem. This design enables pre-computed latent caching by avoiding the prohibitive storage overhead of high-dimensional latents. Furthermore, our specialized decoder is explicitly designed to operate on highly compressed, low-dimensional latents, thereby effectively bridging the gap between semantic-rich VFM features and pixel-accurate image synthesis.
\end{itemize}

%% file: sec_camera_ready/X_suppl_2_cknna.tex
\section{Representation alignment metrics}
\label{appendix:alignment-metrics}

This section outlines the alignment metrics used in our analysis, all aimed at characterizing how two representations maintain local geometric structure. We 
detail CKNNA, which measures agreement in neighborhood relations, and SE-CKNNA, which extends this evaluation to semantic-preserving perturbations for a distribution-aware assessment of alignment stability.

\subsection{CKNNA formulation}
\label{appendix:cknna}

CKNNA~\cite{platonic} (Centered Kernel Nearest-Neighbor Alignment) is a locality-sensitive variant of CKA~\cite{cka} (Centered Kernel Alignment). While CKA measures global similarity between two representations using centered kernel matrices, CKNNA focuses on \textbf{local neighborhood geometry} by restricting the comparison to mutual $k$-nearest-neighbor pairs. This makes it more suitable for evaluating whether representation spaces preserve fine-grained geometric structure.

Formally, given kernel similarity matrices
$K,L\in\mathbb{R}^{n\times n}$ computed from two representations of the same $n$
samples, we construct a mutual $k$-nearest-neighbor mask:

\begin{equation}
A_{ij}=\mathbf{1}\big\{
i\neq j,\ 
j\in\operatorname{knn}_k^{K}(i)\ \wedge\
j\in\operatorname{knn}_k^{L}(i)
\big\}.
\label{eq:supp_mask_def}
\end{equation}

The locally masked kernels are:

\begin{equation}
K^{(k)} = K \odot A,\qquad
L^{(k)} = L \odot A.
\label{eq:supp_masked_kernels}
\end{equation}

Each masked kernel is then double-centered with
$H=I-\tfrac{1}{n}\mathbf{1}\mathbf{1}^\top$:

\begin{equation}
\widetilde{K}=H K^{(k)} H,\qquad
\widetilde{L}=H L^{(k)} H.
\label{eq:supp_center}
\end{equation}

Flattening and $L_2$-normalizing gives:

\begin{equation}
u=\frac{\mathrm{vec}(\widetilde{K})}{\|\widetilde{K}\|_F},\qquad
v=\frac{\mathrm{vec}(\widetilde{L})}{\|\widetilde{L}\|_F}.
\label{eq:supp_unit_vecs}
\end{equation}

The CKNNA score is the normalized inner product:

\begin{equation}
\mathrm{CKNNA}(X)=u^\top v \in [0,1].
\label{eq:supp_cknna}
\end{equation}

Since $\|u\|=\|v\|=1$, this is equivalently:

\begin{equation}
u^\top v
=
1-\tfrac12\|u-v\|_2^2,
\label{eq:supp_cknna_dist}
\end{equation}

\noindent meaning it captures the discrepancy between the masked and centered local kernels of the two representations. Higher CKNNA values indicate better 
preservation of local neighborhood structure and therefore stronger alignment between the two representation spaces.

\subsection{Semantic-Equivariant CKNNA formulation}
\label{appendix:se-cknna}

SE-CKNNA (short for \textbf{S}emantic-\textbf{E}quivariant CKNNA) extends CKNNA to a \textbf{semantic-preserving perturbation distribution}. 
Rather than evaluating alignment only on clean images, it measures CKNNA under semantic-preserving transformations and aggregates the results. We provide visualized example of these preserving transformations on Fig.~\ref{fig:brittleness-detailed}.
Thus, SE-CKNNA is a distribution-aware enhancement of CKNNA, which is \textbf{not} a brand-new metric, but inherits all assumptions and limitations of the original formulation.

Let $\mathcal{T}$ denote a set of semantic-preserving transformations. In practice, these are uniformly sampled from:
\begin{itemize}[leftmargin=*,itemsep=0mm]
    \setlength{\itemsep}{2pt}
    \item additive noise (added in the $[0,1]$ pixel range) with strengths $\{0.05, 0.10, 0.15, 0.20\}$,
    \item scale interpolations $\{0.25, 0.50, 0.75, 1.0\}$,
    \item discrete rotations $\{0^\circ, 90^\circ, 180^\circ, 270^\circ\}$.
\end{itemize}

Noise introduces mild pixel-level perturbation within the normalized $[0,1]$ 
range, while scaling and rotation modify only spatial arrangement and are treated 
jointly as equivariant transformations.

For each $T\in\mathcal{T}$, CKNNA is computed as:

\begin{equation}
\mathrm{CKNNA}(T)=u_T^\top v_T,
\label{eq:supp_cknna_T}
\end{equation}

\noindent where $u_T$ and $v_T$ are the unit-norm vectorizations of the masked and 
double-centered kernels derived from the transformed features 
$\{F(Tx_i)\}_{i=1}^n$.

Since $u_T$ and $v_T$ lie on the unit sphere:
\begin{equation}
u_T^\top v_T
=
1-\frac12\|u_T - v_T\|_2^2,
\label{eq:supp_inner_T}
\end{equation}
\noindent showing that CKNNA($T$) measures the discrepancy between the transformed local kernels of the representations.

SE-CKNNA is defined as the expected CKNNA over a perturbation distribution $p(T)$:
\begin{equation}
\mathrm{SE-CKNNA}
=
\mathbb{E}_{T\sim p(T)}[u_T^\top v_T].
\label{eq:supp_se_expectation}
\end{equation}

Using Eq.~\eqref{eq:supp_inner_T}, this can be written as:
\begin{equation}
\mathrm{SE-CKNNA}
=
1 - \frac12\,\mathbb{E}_{T\sim p(T)}\!\left[\|u_T - v_T\|_2^2\right],
\label{eq:supp_se_distance}
\end{equation}

\noindent clarifying that SE-CKNNA reflects the \textbf{average discrepancy} between the 
masked, centered local kernels induced by different semantic-preserving transformations. In practice, SE-CKNNA is the averaged CKNNA score across sampled transformations, and the relative deviation between SE-CKNNA and clean-image CKNNA reveals the stability of alignment under semantic-preserving perturbations.

\subsection{Evaluation setup}

Following~\cite{platonic}, we use $\mathrm{top}\text{-}k=10$ with channel-wise
normalization and outlier filtering before computing CKNNA. For layer-wise analysis in generative models, we use spatial tokens only (discarding [CLS] tokens as in REG~\cite{reg}) and compute alignment after global pooling along the spatial dimension for both model and reference features. All generative models are evaluated in a unified setting, using the noised latent at diffusion timestep $t=0.5$ with null-conditioning, consistent with the protocol in REPA~\cite{repa}.

%% file: sec_camera_ready/X_suppl_3_arch.tex
\section{Architectural details of VFM-VAE}
\label{appendix:arch}

This section provides implementation-level details of the VFM-VAE architecture that are omitted from the main paper for clarity. We describe the projection module used for latent compression and expansion, the structure of the progressive reconstruction blocks in the decoder, the upsampling unit, and the two-branch latent routing strategy. These components together form the complete tokenizer and decoder design used in our experiments.

\begin{figure*}[ht]
  \centering
  \includegraphics[width=0.9\linewidth]{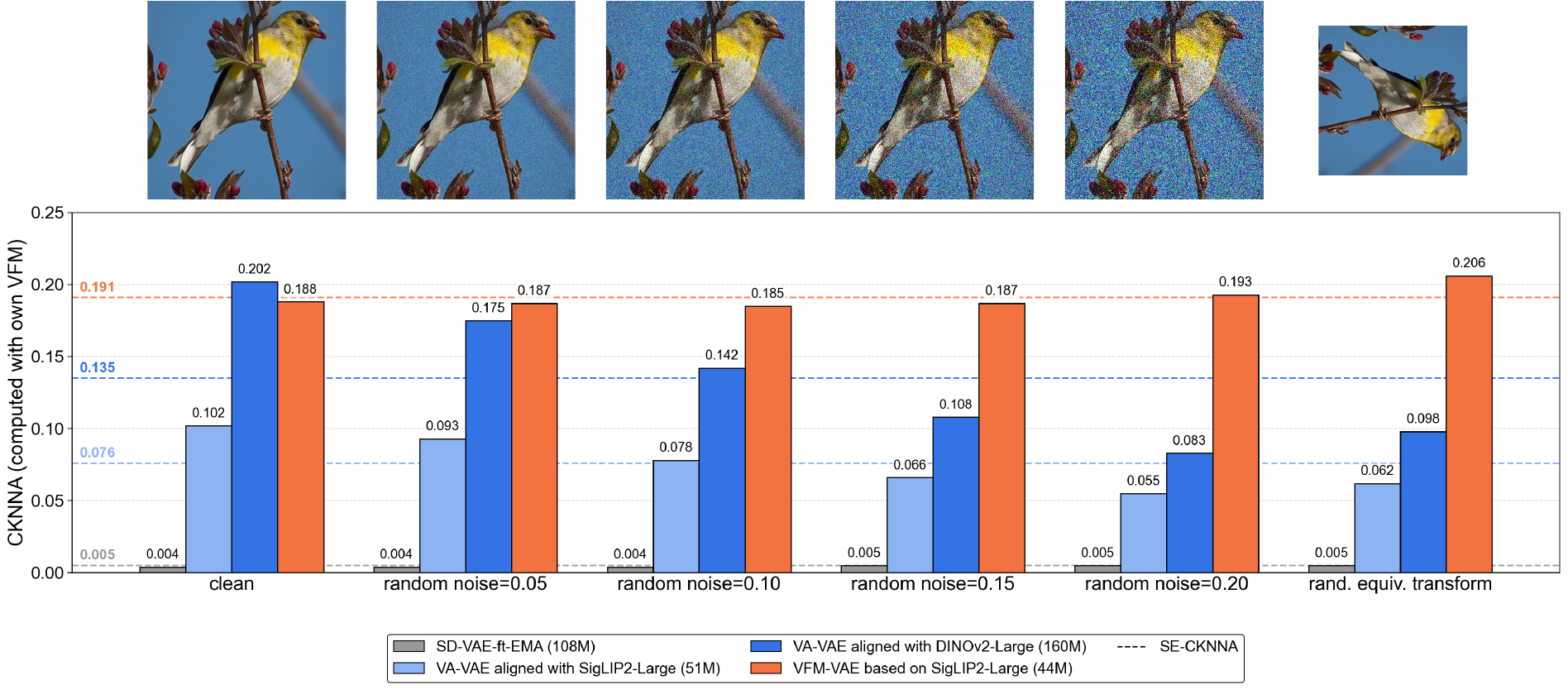}
   \caption{Brittleness of aligned representations under semantic-preserving transformations. Specifically, CKNNA~\cite{platonic} value for SD-VAE~\cite{ldm} is computed with DINOv2-Large~\cite{dinov2}. \emph{M} denotes the number of the training images in millions. Under semantic-preserving transformations, VFM-VAE demonstrates notably stronger alignment with VFM features than all VA-VAE variants and SD-VAE.}
  \label{fig:brittleness-detailed}
\end{figure*}

\begin{table*}[ht]
\centering
\caption{Ablation of VFM-VAE components. Modules are added while maintaining light-weight alignment to the VFM.}
\label{tab:detailed-ablation}
\begin{tabular}{@{}l|c|ccccc@{}}
\toprule
Setting                      & \ Trainable params & rFID$\downarrow$ & rIS$\uparrow$ & LPIPS$\downarrow$ & PSNR$\uparrow$ & SSIM$\uparrow$ \\ \midrule
SD-VAE-style Baseline        & 43.0M    & 19.69 & 74.9 & 0.456     & 14.59    & 0.264    \\
+ Multi-scale Latent Fusion  & 88.0M    & 14.35 & 93.6 & 0.433     & 14.71    & 0.241    \\
+ Our Modern Blocks          & 132.3M   & 1.08 & 194.6 & 0.291     & 18.06    & 0.388    \\
+ Encoder Modifications      & 140.6M   & \textbf{0.71} &  \textbf{206.8} & \textbf{0.202}     & \textbf{22.54}    & \textbf{0.571}    \\
Scaled SD-VAE-style Baseline & 150.9M   & 38.46    & 40.3   & 0.549     & 13.59    & 0.208    \\ \bottomrule
\end{tabular}
\end{table*}

\begin{table*}[ht]
\centering
\caption{Separate ablation of the global pooled feature. Modules are added while maintaining light-weight alignment to the VFM.}
\label{tab:ablation-global}
\begin{tabular}{l|ccccc}
\hline
Method & gFID$\downarrow$          & gIS$\uparrow$             & rFID$\downarrow$ & rIS$\uparrow$          & rPSNR$\uparrow$            \\ \hline
w/o global feature & 10.51         & 96.44    & 0.72 & \textbf{208.86}          & \textbf{22.56}       \\
with global feature  & \textbf{9.88} & \textbf{99.05} & \textbf{0.71} & 206.80 & 22.54          \\ \hline
\end{tabular}
\end{table*}

\begin{table}[t]
\centering
\caption{VFM-VAE's reconstruction metrics across 256 and 512 resolutions, evaluated on ImageNet validation set.}
\label{tab:detailed-recon}
\begin{tabular}{@{}l|ccccc@{}}
\toprule
Resolution & rFID & rIS   & LPIPS & PSNR  & SSIM  \\ \midrule
256x256        & 0.52 & 214.1 & 0.221 & 22.99 & 0.593 \\
512x512        & 0.44 & 235.8 & 0.210 & 25.52 & 0.682 \\ \bottomrule
\end{tabular}
\end{table}

\subsection{Attention Projection}

\noindent \textbf{Motivation.} We adopt the \emph{Attention Projection} module from UniTok~\cite{unitok} for its simplicity, stability, and computational efficiency. It provides a unified mechanism for channel compression before sampling and decompression during decoding, maintaining a well-structured representation distribution that remains aligned with VFM features.

\noindent \textbf{Encoder-side (compression).}
We extract shallow, middle, and final features from the frozen VFM. 
If their spatial sizes (and thus tokenizations) do not match the target latent configuration, we first apply \texttt{PixelShuffle}~\cite{pixelshuffle} to reconcile the spatial mismatch and then concatenate the tokens. 
\textbf{Importantly, the concatenated tokens are passed only once through the \emph{Attention Projection}} to obtain the latent representation, from which we compute distribution statistics (e.g., mean and variance) and sample the latent code.

\noindent  \textbf{Decoder-side (decompression).}
During decoding, \emph{Attention Projection} is used to expand channel dimensions and distribute the outputs to the semantic and spatial branches of the decoder, allowing the model to recombine abundant low-level cues for high-quality reconstruction.

\subsection{Global and spatial branches}

The decoder receives two complementary latent streams: a \textbf{global branch} and a \textbf{spatial branch}.  
The global branch is a lightweight MLP mapping network that transforms the pooled latent vector into a global style latent feature, providing semantic control and ensuring consistent appearance across scales.  
The spatial branch processes multi-resolution latent features through hierarchical convolutional blocks. Each latent is adapted to the target resolution via \texttt{PixelUnshuffle} or \texttt{PixelShuffle}, combined with $3{\times}3$ and $1{\times}1$ convolutions, \texttt{GroupNorm}, and \texttt{GELU} activation.  
Together, the two branches fuse holistic semantics with fine-grained spatial details, enabling the decoder to generate coherent and detailed reconstructions across progressive resolutions.

\subsection{Progressive reconstruction blocks}

Each reconstruction block $\mathcal{B}_i$ is responsible for processing and refining features at its designated spatial resolution. Let $\mathbf{h}^{(i)}$ denote the feature map at stage $i$, and let $\mathbf{z}_g$ and $\mathbf{z}_s^{(i)}$ be the global and spatial latents. The decoder updates $\mathbf{h}^{(i)}$ using the following stage-dependent formulation:
\begin{equation}
\mathbf{h}^{(i)} =
\begin{cases}
\mathcal{B}_i(\mathrm{Up}(\mathbf{z}_s^{(1)}),\, \mathbf{z}_g), & i = 1, \\[4pt]
\mathcal{B}_i(\mathrm{Up}(\mathrm{Concat}[\mathbf{h}^{(i-1)},\, \mathbf{z}_s^{(i)}]),\, \mathbf{z}_g), & 2 \le i \le 4, \\[4pt]
\mathcal{B}_i(\mathrm{Up}(\mathbf{h}^{(i-1)}),\, \mathbf{z}_g), & 5 \le i \le 6,
\end{cases}
\label{eq:modulated_block}
\end{equation}
where $\mathrm{Up}(\cdot)$ denotes a $2\times$ spatial upsampling.  
Spatial latents $\mathbf{z}_s^{(i)}$ are injected only in the first four stages, where coarse and mid-level structures are reconstructed. The final stages operate solely on the progressively refined features $\mathbf{h}^{(i-1)}$ under global modulation from $\mathbf{z}_g$, focusing on high-frequency detail synthesis.

\subsection{Upsampling modules}

We improve the StyleGAN-T~\cite{stylegan-t} upsampling unit with PyTorch implementation for better readability and extensibility. The module normalizes input features via \texttt{GroupNorm}, applies $3\times3$ depthwise and $1\times1$ pointwise convolutions for local extraction and channel mixing, upsamples with \texttt{PixelShuffle}, and finally applies a fixed Gaussian blur to suppress checkerboard artifacts. It serves two roles: (1) progressively increasing spatial resolution 
across backbone blocks, and (2) refining features in the output pathway before the ToRGB head. This design retains StyleGAN-T’s efficiency while improving stability and visual fidelity.

%% file: sec_camera_ready/X_suppl_4_analysis.tex
\begin{table*}[t]
\centering
\caption{Comparison of reconstruction and generation quality under fair setting of VFM and training duration.
* denotes the reproduced VA-VAE by us with its open-sourced code.
CKNNA is computed with each respective VFM. Details of CKNNA and SE-CKNNA are provided in \cref{appendix:alignment}. \textit{Relative Change} represents the variation between SE-CKNNA and CKNNA, calculated as $|\text{SE-CKNNA} - \text{CKNNA}| / \text{CKNNA}$. Generation metrics are reported without CFG~\cite{cfg}.}
\label{tab:vfm-fair-comp}
\resizebox{\textwidth}{!}{%
\begin{tabular}{lccccc|cc|cc}
\toprule
\multirow{2}{*}{Tokenizer} & \multirow{2}{*}{VFM} & \multirow{2}{*}{Training Duration} & \multirow{2}{*}{CKNNA} & \multirow{2}{*}{SE-CKNNA} & \multirow{2}{*}{Relative Change}
& \multicolumn{2}{c|}{\textit{Reconstruction}} & \multicolumn{2}{c}{\textit{Generation}} \\
 &  &  &  &  &  & rFID$\downarrow$ & rIS$\uparrow$ & gFID$\downarrow$ & gIS$\uparrow$ \\
\midrule
VA-VAE   & DINOv2-Large   & 160M (125 epochs)   & 0.202 & 0.135 & -33.2\% & \textbf{0.30} & 213.6 & 5.14 & 130.2 \\
VA-VAE*   & SigLIP2-Large  & 51M (40 epochs)   & 0.102 & 0.076 & -25.5\% & 0.84 & 207.4 & 7.83 & 115.1 \\
VFM-VAE  & SigLIP2-Large  & 44M ($\approx$ 38 epochs) & 0.188 & 0.191 & +1.6\% & 0.52 & \textbf{214.1} & \textbf{3.80} & \textbf{152.8} \\
\bottomrule
\end{tabular}%
}
\end{table*}


\begin{table*}[ht]
\centering
\caption{Comparison of VFM-VAE variants across different VFMs. Each tokenizer is trained through two-stage alignment (5M strong/weak). Generation results are reported at 100k training steps using LightningDiT-L/1 without CFG.}
\label{tab:ablation-vfm}
\resizebox{0.87\textwidth}{!}{%
\begin{tabular}{@{}l|ccccc|ccccc@{}}
\toprule
\multirow{2}{*}{VFM} & \multicolumn{5}{c|}{\textit{Reconstruction}}                                               & \multicolumn{5}{c}{\textit{Generation}}                                                 \\
                     & rFID$\downarrow$          & rIS$\uparrow$             & LPIPS$\downarrow$           & PSNR$\uparrow$            & SSIM$\uparrow$            & gFID$\downarrow$           & sFID$\downarrow$           & gIS$\uparrow$             & Precision$\uparrow$     & Recall$\uparrow$        \\ \midrule
EVA-CLIP-Large       & \textbf{1.35} & 188.4          & \textbf{0.300} & \textbf{19.33} & \textbf{0.431} & 4.40          & 5.13          & 146.4          & \textbf{0.80} & \textbf{0.58} \\
DINOv2-Large         & 1.55          & \textbf{199.8} & 0.329          & 17.60          & 0.362          & \textbf{4.00} & 5.16          & \textbf{147.1} & \textbf{0.80} & \textbf{0.58} \\
SigLIP2-Large        & 1.61          & 178.0          & 0.322          & 18.73          & 0.408          & 5.59          & \textbf{4.85} & 127.8          & 0.79          & 0.57          \\ \bottomrule
\end{tabular}%
}
\end{table*}

\begin{table*}[ht]
\centering
\caption{Comparison of ImageNet-512 generation performance after 100k training steps without CFG using LightningDiT-B/1. VFM-VAE demonstrates improved generation quality over the VA-VAE baseline.}
\label{tab:higher-resolution}
\resizebox{0.6\textwidth}{!}{%
\begin{tabular}{@{}l|ccccc@{}}
\toprule
Method                     & gFID$\downarrow$ & sFID$\downarrow$ & gIS$\uparrow$ & Precision$\uparrow$ & Recall$\uparrow$ \\ \midrule
VA-VAE + LightningDiT-B/1  & 21.42    & \textbf{5.65}    & 55.3   & 0.75     & \textbf{0.60}    \\
VFM-VAE + LightningDiT-B/1 & \textbf{18.05}    & 7.11    & \textbf{69.6}   & \textbf{0.78}     & \textbf{0.60}    \\ \bottomrule
\end{tabular}%
}
\end{table*}

\section{More ablation studies of VFM-VAE}
\label{appendix:analysis}

In this section, we provide additional analyses of VFM-VAE, including detailed reconstruction and alignment results, its compatibility with different VFMs, comparison with VA-VAE fair baseline, higher-resolution generation, and text-to-image generation.

\subsection{Detailed reconstruction metrics}

To comprehensively evaluate the reconstruction capability of VFM-VAE, we report a full set of perceptual and pixel-level metrics, including LPIPS~\cite{lpips}, PSNR, and SSIM, on both ImageNet-256 and ImageNet-512 validation set, as shown in~\cref{tab:detailed-recon}. The 512-resolution model is obtained by fine-tuning from the 256-resolution pre-trained weights (details in~\cref{appendix:higher-resolution}). Benefiting from the architectural designs introduced in \cref{appendix:arch} and the main paper Sec.~3, VFM-VAE achieves considerable reconstruction performance across all metrics at both resolutions.


In addition, we provide more detailed ablation studies in~\cref{tab:detailed-ablation}. We attempt to scale a SD-VAE-like baseline to match the trainable parameter count of VFM-VAE. However, this leads to unstable or even collapsed training, ultimately resulting in worse reconstruction quality. 
Furthermore, in~\cref{tab:ablation-global}, we conduct a separate ablation on the global pooled feature design; removing it has almost no impact on reconstruction but weakens the generation performance. 
These results indicate that the additional parameters introduced in VFM-VAE together with its modular design are effective and more stable to train than SD-VAE original design, delivering consistent improvements across all reconstruction metrics.

\subsection{Detailed representation alignment metrics}
\label{appendix:alignment}

In~\cref{fig:brittleness-detailed}, we further present a detailed CKNNA-based analysis under small image perturbations. The example images above illustrate the types of applied transformations, demonstrating that the semantic content remains intact. Besides the original VA-VAE~\cite{vavae} (aligned with DINOv2-Large~\cite{dinov2}), we also train a fair VA-VAE variant that is aligned with the same SigLIP2~\cite{siglip2} backbone as VFM-VAE and uses a comparable amount of training data. Even under these matched conditions, VA-VAE still fails to achieve robust alignment. In contrast, VFM-VAE, which employs a frozen VFM encoder as its front-end, exhibits significantly stronger representation alignment performance.

\subsection{Not only one VFM choice}
\label{appendix:vfm-choice}
We evaluate VFM-VAE's compatibility with three representative VFMs: EVA-CLIP-Large~\cite{eva-clip}, DINOv2-Large~\cite{dinov2}, and SigLIP2-Large~\cite{siglip2}. In detail, DINOv2-Large is a self-supervised vision model trained exclusively on image data, providing robust semantic representations without text alignment. EVA-CLIP-Large and SigLIP2-Large are VFMs trained on large-scale image-text pairs. While EVA-CLIP continue the original CLIP's contrastive learning approach to scale up, SigLIP2 employs a combination of image-language contrastive loss, caption prediction loss, and self-supervised loss to produce dense visual representations.


Under the same training schedule, we report both reconstruction and generation metrics in \cref{tab:ablation-vfm}. In terms of reconstruction, EVA-CLIP-Large achieves the best performance across most metrics. While DINOv2-Large slightly outperforms SigLIP2-Large in realism-oriented metrics (rFID and rIS), it falls significantly short in perceptual (LPIPS) and pixel-level (PSNR and SSIM) metrics. Interestingly, in terms of generation performance, DINOv2 leads the field, followed by EVA-CLIP-Large, with SigLIP2-Large ranking last. This superiority likely stems from DINOv2’s purely image-centric self-supervised pre-training, which makes it more compatible with nearly pure image generation tasks like ImageNet, where only class information is introduced.

Despite these findings, considering SigLIP2-Large's diverse training objectives, it provides more robust downstream performance and native text-alignment capabilities, which are crucial for text-to-image (T2I) generation (see \cref{appendix:vlm-t2i}). We therefore adopt SigLIP2-Large as our default VFM for long-term training to balance reconstruction, alignment, and T2I performance. Even though SigLIP2 is not the optimal choice for an object-oriented dataset like ImageNet, we still achieve impressive results in both reconstruction and generation. These findings demonstrate that VFM-VAE maintains strong performance across diverse VFM families, confirming its architectural generality rather than a dependence on any specific model.

The differing feature emphases of the DINOv2\mbox{-}Large and SigLIP2\mbox{-}Large variants of VFM\mbox{-}VAE, together with the performance gap observed between their corresponding VFM-VAE variants, particularly the weaker pixel-level reconstruction of the DINOv2-Large variant, naturally raise a key question: \textit{Are the two observed gaps, the tokenizer’s alignment gap and the downstream diffusion generation gap between VFM\mbox{-}VAE based on SigLIP2\mbox{-}Large and VA\mbox{-}VAE aligned with DINOv2\mbox{-}Large, simply consequences of differences in the underlying VFMs?} We address this issue in \cref{appendix:fair}.

\subsection{Improvement is not due to a different VFM}
\label{appendix:fair}

VFM-VAE consistently achieves faster convergence and higher generation quality than the VA-VAE baseline across all generative models.
A key distinction, however, lies in the choice of underlying VFMs for alignment: VFM-VAE is built upon SigLIP2-Large, trained primarily with contrastive objectives to support both vision-language alignment and dense visual representation learning, whereas VA-VAE is aligned with DINOv2-Large, a purely vision-based self-supervised model.
This difference introduces a natural trade-off between alignment and reconstruction quality.
As shown in \cref{appendix:vfm-choice}, when VFM-VAE is aligned with DINOv2-Large, its reconstruction slightly lags behind that of the SigLIP2-Large variant.

To determine whether the performance gap originates from the VFMs themselves, we conducted a controlled comparison.
Following VA-VAE’s strong alignment setup, we retrained a VA-VAE using \textbf{SigLIP2-Large on 51M images ($\approx$ 40 epochs)}, achieving adequate reconstruction performance before training the same generative model.


Combining \cref{fig:brittleness-detailed} and \cref{tab:vfm-fair-comp}, two observations emerge:

\begin{itemize}[leftmargin=*,itemsep=0mm]

\item In terms of the tokenizer’s intrinsic alignment and reconstruction capability, VA-VAE requires substantially more training images to achieve competitive alignment and reconstruction performance, whereas VFM-VAE attains a more favorable balance with significantly fewer training images and exhibits more robust alignment to its underlying VFM. Even when using the same VFM, the VA-VAE variant still experiences a notable drop in alignment under small input perturbations.

\item In terms of generative performance, even when the VFM choice, the number of training images, and the diffusion-training configuration are matched, the VA-VAE system still lags significantly behind the VFM-VAE system.

\end{itemize}

In summary, the performance advantage of VFM-VAE does not arise merely from employing a stronger VFM. Instead, it stems from its architectural design, which leverages frozen VFM features as the starting point, enabling stronger and more stable representation learning the generative learning through the whole pipeline.

\subsection{Preliminary generation on 512$\times$512 resolution}
\label{appendix:higher-resolution}

Limited by resources, we present ablation studies comparing our tokenizer with VA-VAE on reduced model sizes. We evaluate VFM-VAE on ImageNet-512~\cite{imagenet}.

\noindent \textbf{Setting.}
To maintain continuity with the 256-resolution setup, we reuse the same VFM configuration (SigLIP2-Large-Patch16-512~\cite{siglip2}) and all corresponding model components, and fine-tune them jointly at 512 resolution.
The main difference from the 256-resolution setup lies in how images are fed into the VFM: while 256-resolution images were first upsampled 2$\times$ before being passed into the VFM, the 512-resolution images are directly fed into the VFM, which avoids interpolation overhead and reduces memory consumption.

To enable efficient fine-tuning from the 256-resolution model, we set the output dimension of the “encoded features” in \cref{tab:vfmvae-struct-hyperparams} to 256, and retrain only the Attention Projection module that extracts multi-layer VFM features. All remaining modules reuse the pretrained weights from the 256-resolution model. The training hyperparameters follow the first three stages in \cref{tab:vfmvae-train-hyperparams}; we only adjust the number of fine-tuning images for the strong-alignment, weak-alignment, and SSIM phases to 1M, 500k, and 500k respectively, while keeping the multi-scale pixel loss enabled throughout. The reconstruction results after 512-resolution fine-tuning are reported in \cref{tab:detailed-recon}.

\noindent \textbf{Results.}
Finally, we train both VA-VAE and VFM-VAE with LightningDiT-B/1 for 100k steps (80 epochs) on ImageNet-512 and report generation performance without CFG~\cite{cfg}. As shown in \cref{tab:higher-resolution}, the VFM-VAE system achieves clearly superior generative performance compared with the VA-VAE system.

\subsection{Text-to-image generation}
\label{appendix:vlm-t2i}

\noindent \textbf{Motivation.}
While the main paper focuses on class-based image generation, it remains important to examine how VFM-VAE performs when integrated into text-to-image generation and multimodal systems.  
A strong visual tokenizer should not only reconstruct accurately but also provide semantically consistent latents that interface smoothly with Vision-Language Models (VLMs).  
To validate this, we combine VFM-VAE and VA-VAE respectively with \textbf{BLIP3-o}~\cite{blip3o} in a unified text-to-image framework and compare their effectiveness in generative modeling.
Due to limited computing resources, here we present an ablation comparing our tokenizer with VA-VAE on limited training steps.

\noindent \textbf{Setting.}
Given a text prompt, BLIP3-o first produces a fixed number of tokens (default 64).  
We extract the hidden states before the LM head as semantic conditioning for the diffusion model.  
During training, the diffusion model predicts a latent that is \emph{flow-matched}~\cite{flow} to the VAE encoder latent; during inference, this latent is decoded by the VAE decoder to generate the final image with 256 resolution.
The visual–language backbone is \textbf{Qwen2.5-VL-3B-Instruct}~\cite{qwen_2_5_vl}, and the diffusion backbone is Lumina-Next (DiT)~\cite{lumina}, where the patch size is reduced to 1 and the input/output channels are aligned to a latent of $16\times32\times32$.  
We pretrain the system for \textbf{one epoch} on the official BLIP3-o pretraining corpora, focusing solely on the text-to-image objective.  
Since more than 80\% of the dataset consists of long prompts, we omit GenEval and evaluate exclusively on DPG-Bench (higher score is better)~\cite{dpgbench} and MJHQ-30K (lower gFID is better)~\cite{mjhq30k}.

\noindent \textbf{DPG-Bench results.}
VFM-VAE + BLIP3-o achieves a higher overall score (59.1) than VA-VAE + BLIP3-o (55.4) (see \cref{tab:dpgbench}).  
At the L1 level, improvements are evident in \emph{relation}, \emph{global}, and \emph{other} categories, while a slight decrease is observed in \emph{entity}.  
This indicates that VFM-VAE provides stronger global and relational understanding, enhancing compositional reasoning and text–image alignment, albeit with slightly weaker recall of local structures.

\noindent \textbf{MJHQ-30K results.}
VFM-VAE + BLIP3-o also achieves significantly lower gFID across nearly all categories (see 
\cref{tab:mjhq30k}), reducing the overall score from 23.00 to 16.98.  
Notable gains are seen in \emph{animals} $(44.78 \rightarrow 32.08)$, \emph{fashion} $(42.80 \rightarrow 30.27)$, \emph{indoor} $(44.01 \rightarrow 34.37)$, and \emph{people} $(48.65 \rightarrow 36.62)$, with comparable results on \emph{plants} and a minor trade-off in \emph{logo}.

\noindent \textbf{Summary.}
Under identical VLM and diffusion backbones, replacing VA-VAE with VFM-VAE yields more semantically aligned and generation-friendly latents, leading to higher text–image consistency, and overall better visual quality—even with only one epoch of pretraining.

\begin{table*}[t]
\centering
\caption{Text-to-image generation results with BLIP3-o on DPG-Bench (higher score is better, evaluated at 256 resolution after 1 epoch of pretraining). 
VFM-VAE + BLIP3-o achieves higher overall scores, indicating stronger text–image alignment.}
\label{tab:dpgbench}
\setlength{\tabcolsep}{5pt}
\resizebox{0.6\linewidth}{!}{
\begin{tabular}{@{}l|ccccc|c@{}}
\toprule
Text-to-image Model & entity & global & other & attribute & relation & overall \\
\midrule
VA-VAE + BLIP3-o   & \textbf{73.2} & 69.1 & 71.2 & 70.1 & 77.9 & 55.4 \\
VFM-VAE + BLIP3-o  & 68.4 & \textbf{70.9} & \textbf{75.2} & \textbf{71.1} & \textbf{80.0} & \textbf{59.1} \\
\bottomrule
\end{tabular}
}
\end{table*}

\begin{table*}[t]
\centering
\caption{Text-to-image generation results with BLIP3-o on MJHQ-30K (lower gFID is better, evaluated at 256 resolution after 1 epoch of pretraining). 
VFM-VAE + BLIP3-o achieves significantly lower gFID.}
\label{tab:mjhq30k}
\resizebox{\textwidth}{!}{
\begin{tabular}{@{}l|ccccccccccc@{}}
\toprule
Text-to-image Model & animals & art & fashion & food & indoor & landscape & logo & people & plants & vehicles & overall \\ 
\midrule
VA-VAE + BLIP3-o  & 44.8 & 43.4 & 42.8 & 46.1 & 44.0 & 47.4 & \textbf{59.0} & 48.7 & \textbf{50.7} & 40.5 & 23.0 \\
VFM-VAE + BLIP3-o & \textbf{32.1} & \textbf{36.0} & \textbf{30.3} & \textbf{44.7} & \textbf{34.4} & \textbf{41.3} & 60.4 & \textbf{36.6} & 50.9 & \textbf{39.2} & \textbf{17.0} \\
\bottomrule
\end{tabular}
}
\end{table*}

%% file: sec_camera_ready/X_suppl_5_implementation.tex
\section{Implementation details}
\label{appendix:details}


\textbf{VFM-VAE training.} Model hyperparameters are provided in \cref{tab:vfmvae-struct-hyperparams}, with the stable training recipe detailed in \cref{tab:vfmvae-train-hyperparams}. We find that training stability is preserved under moderate weight adjustments, provided that the adversarial loss remains stable. Our multi-stage training strategy follows the general structure of VA-VAE~\cite{vavae}. In the strong alignment stage, large representation regularization losses are applied to quickly establish VFM–VAE alignment. In the weak alignment stage, the weight of this loss is reduced to maintain alignment while shifting focus toward reconstruction quality. Notably, VA-VAE is trained on the full ImageNet training set (1,281,167 images), whereas VFM-VAE is trained on a filtered subset containing only images with a minimum resolution of 256 (1,152,196 images). We further introduce two fine-tunings: 

\begin{itemize}[leftmargin=*,itemsep=0mm]
\item \textbf{SSIM fine-tuning.} Rapid convergence in reconstruction can occasionally cause RGB channel misalignment, leading to color noise near edges. We apply an SSIM loss for targeted refinement.
\item \textbf{PatchGAN~\cite{pix2pix} fine-tuning.} The original DINO~\cite{dino}-based discriminator, due to its large patch size, provides weak supervision for fine details. Adding a finer-grained PatchGAN discriminator improves reconstruction fidelity.
\end{itemize}

The improvements of reconstruction and alignment across the training stages are summarized in \cref{tab:vfmvae-stages}.

\noindent \textbf{Diffusion model training.} LightningDiT follows the same configuration as the VA-VAE system.
When using REG, we apply several adjustments: the latent size changes from $32 \times 32 \times 4$ to $16 \times 16 \times 32$; the SiT-XL~\cite{sit} patch size is reduced from 2 to 1; batch size is increased from 256 to 1024; the learning rate is doubled; $\beta_2$ of AdamW~\cite{adamw} optimizer is reduced from 0.999 to 0.95; and QK normalization~\cite{qknorm} is added to the attention module.
These modifications collectively stabilize long-term training.
All experiments are conducted on a single node with 8$\times$192GB NVIDIA B200 GPUs. In~\cref{ablation-time}, we further measure total wall-clock time to reach equivalent performance. VFM-VAE shows clear speedup over VA-VAE.

\begin{table*}[ht]
\centering
\caption{Reconstruction and alignment performance across 256-resolution training stages of VFM-VAE. CKNNA is computed wtih SigLIP2-Large.}
\label{tab:vfmvae-stages}
\resizebox{0.85\textwidth}{!}{
\begin{tabular}{l | c c c c c c}
\toprule
Training stage & rFID$\downarrow$ & rIS$\uparrow$ & LPIPS$\downarrow$ & PSNR$\uparrow$ & SSIM$\uparrow$ & CKNNA$\uparrow$ \\
\midrule
Stage 1: Strong alignment & 1.05 & 198.2 & 0.266 & 20.39 & 0.473 & \textbf{0.221} \\
Stage 2: + Weak alignment   & 0.60 & 210.3 & \textbf{0.196} & 22.89 & 0.586 & 0.188 \\
Stage 3: + SSIM fine-tuning & 0.54 & 211.2 & 0.209 & 22.85 & 0.586 & 0.188 \\
Stage 4: + PatchGAN fine-tuning & \textbf{0.52} & \textbf{214.1} & 0.221 & \textbf{22.99} & \textbf{0.593} & 0.188 \\
\bottomrule
\end{tabular}}
\end{table*}

\begin{table*}[t]
\centering
\caption{Comparison of generative performance and training costs in A100 GPU hours.}
\label{ablation-time}
\begin{tabular}{lccccc}
\hline
\multirow{2}{*}{Method} & \multirow{2}{*}{gFID$\downarrow$} & \multirow{2}{*}{gIS$\uparrow$} & \multicolumn{3}{c}{A100 GPU hours$\downarrow$} \\ \cline{4-6}
                      &   &   & VAE  & diffusion & total      \\ \hline
VA-VAE + LightningDiT(800 epochs) & 2.17  & 205.6 & 2836 & 3451      & 6287    \\
VFM-VAE + LightningDiT(560 epochs) & \textbf{2.06} & \textbf{205.8} & \textbf{2400} & \textbf{2419}      & \textbf{4819}    \\ \hline
\end{tabular}
\end{table*}

\begin{table*}[ht]
\centering
\caption{VFM-VAE architecture hyperparameters at 256-resolution training.}
\label{tab:vfmvae-struct-hyperparams}
\resizebox{0.7\textwidth}{!}{
\begin{tabular}{c c c}
\toprule
\textbf{Category} & \textbf{Name} & \textbf{Value} \\
\midrule
VFM Backbone & VFM name & SigLIP2-Large-Patch16-512 \\
\midrule
\multirow{2}{*}{Encoded Features} 
& from which layers & [0, 12, -1] \\
& output dims & [64, 64, 64] \\
\midrule
\multirow{4}{*}{Latent} 
& how to compress / decompress & \emph{Attention Projection} \\
& spatial-compression ratio & 16 \\
& latent channels & 32 \\
& channel-decompression ratio & 32 \\
\midrule
\multirow{2}{*}{Spatial Control} 
& block indices & [0, 1, 2, 3] \\
& mapped dims & [512, 256, 128, 128] \\
\midrule
\multirow{2}{*}{Attention} 
& block indices & [0, 1, 2] \\
& attention depths & [2, 2, 2] \\
\bottomrule
\end{tabular}}
\end{table*}

\begin{table*}[ht]
\centering
\caption{VFM-VAE training hyperparameters at 256-resolution training.}
\label{tab:vfmvae-train-hyperparams}
\resizebox{\textwidth}{!}{
\begin{tabular}{c c c c c}
\toprule
\textbf{Setting} & \textbf{Strong Alignment} & \textbf{Weak Alignment} & \textbf{SSIM Fine-tuning} & \textbf{PatchGAN Fine-tuning} \\
\midrule
Batch size & \multicolumn{4}{c}{512} \\
Optimizer & \multicolumn{4}{c}{Adam} \\
Betas & \multicolumn{4}{c}{(0.0, 0.99)} \\
\midrule
Learning rate & $1\times10^{-4}$ & $1\times10^{-4}$ & $5\times10^{-5}$ & $5\times10^{-5}$ \\
L1 loss weight & 1.0 & 1.0 & 1.0 & - \\
LPIPS loss weight & 10.0 & 10.0 & 2.0 & - \\
DINO discriminator loss weight & 1.0 & 1.0 & 1.0 & 1.0 \\
PatchGAN discrminator loss weight & - & - & - & 1.0 \\
Feature matching loss weight~\cite{pix2pix} & - & - & - & 10.0 \\
SSIM loss weight & - & - & 1.0 & - \\
Multiscale pixel loss weight & 0.1 (to 5M = 0) & - & - & - \\
Representation reglarization loss weight & 5.0 & 1.0 & - & - \\
KL loss weight & $1\times10^{-6}$ & $1\times10^{-6}$ & - & - \\
Trainable parameters & Entire tokenizer & Entire tokenizer & The decoder & The second half of the decoder \\
Equivariance regularization~\cite{eqvae} & Yes & Yes & Yes & No \\
Training images & 20M & 20M & 1M & 3M \\
\bottomrule
\end{tabular}}
\end{table*}

%% file: sec_camera_ready/X_suppl_6_qualitation.tex
\section{More qualitative results}
\label{appendix:qualitative}

\begin{figure*}[ht]
    \centering
    \includegraphics[width=\textwidth]{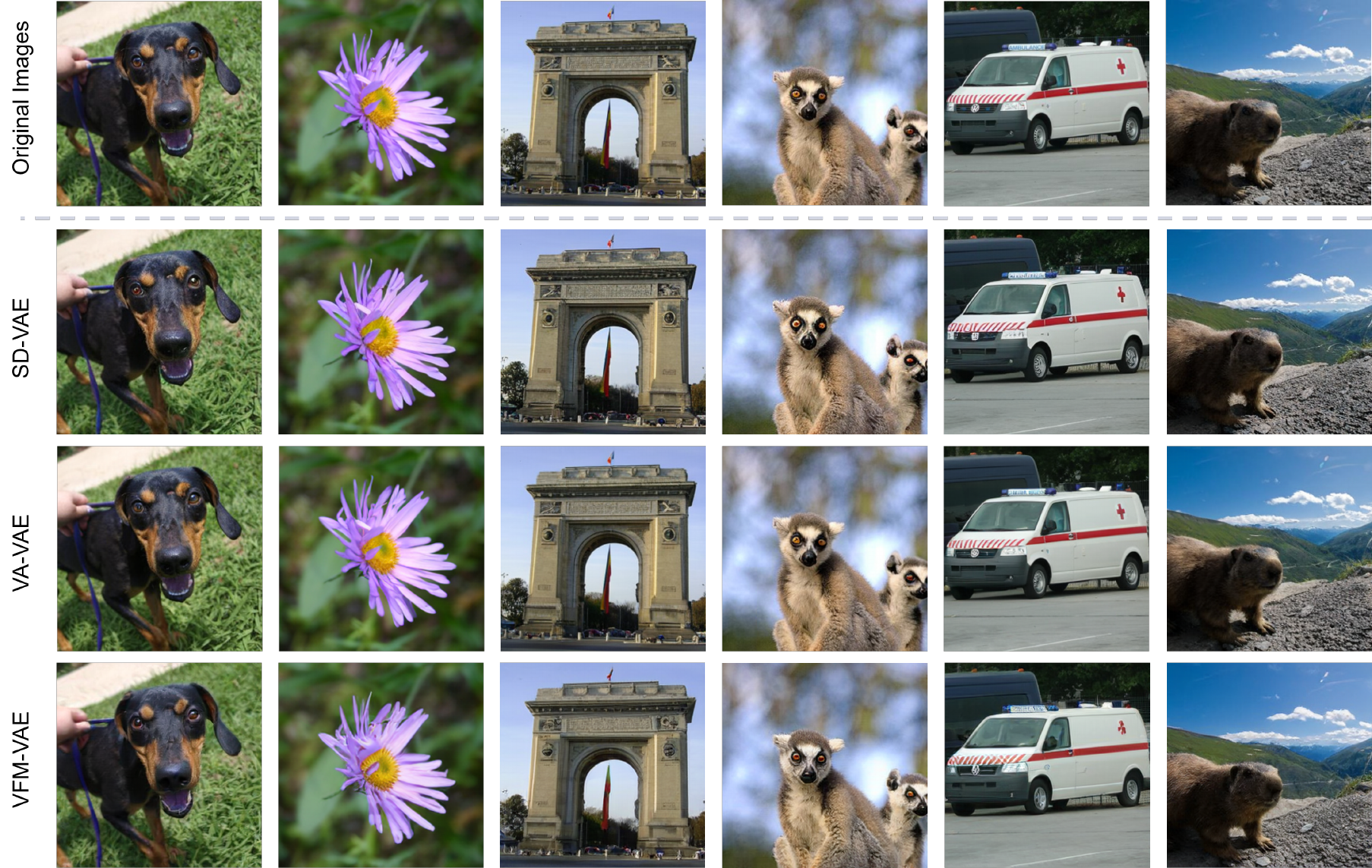}
    \caption{Qualitative comparison of reconstructions from different VAEs.}
    \label{fig:vae_recon}
\end{figure*}

\begin{figure*}[ht]
\centering
\includegraphics[width=\linewidth]{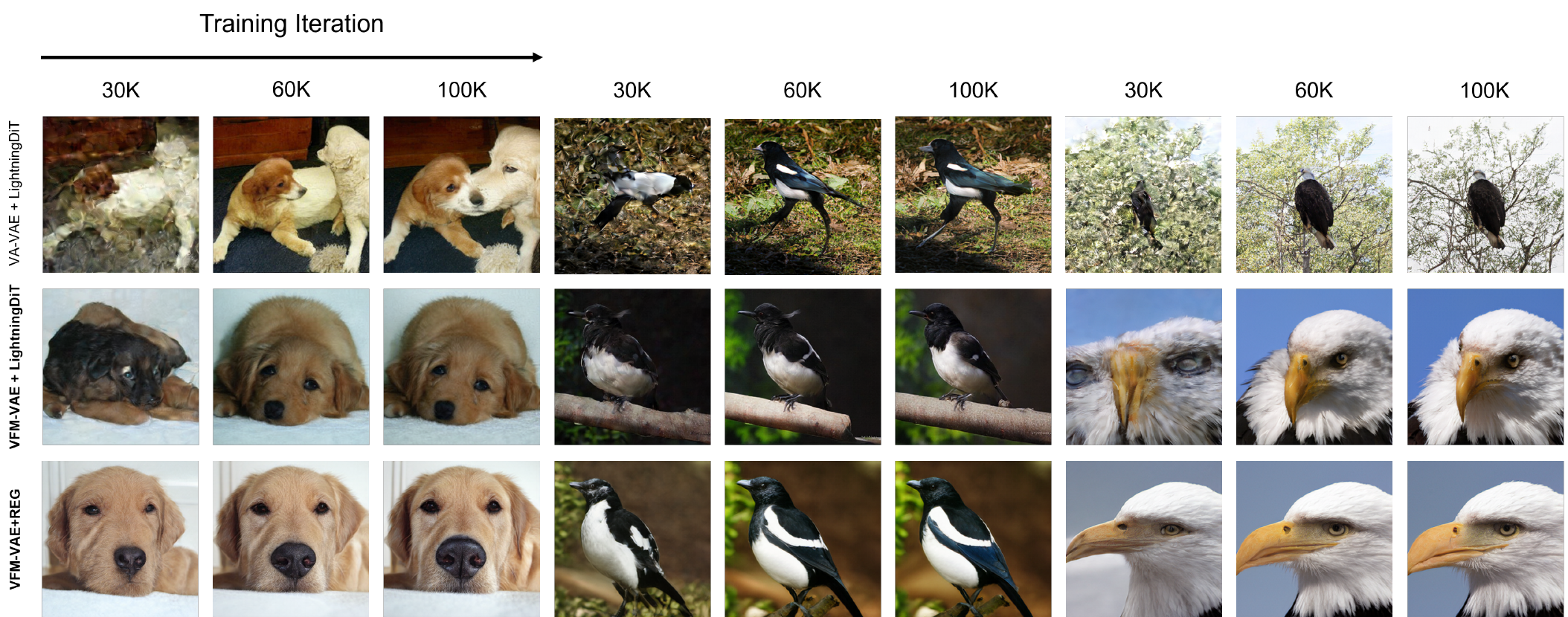}
\caption{Stage-wise visualization of generative model training results. Shown under a fixed random seed and identical initial noise, our approach demonstrates impressive performance and greatly accelerates image generation learning.}
\label{fig:generation}
\end{figure*}

\begin{figure*}[ht]
  \centering
  \includegraphics[width=\textwidth]{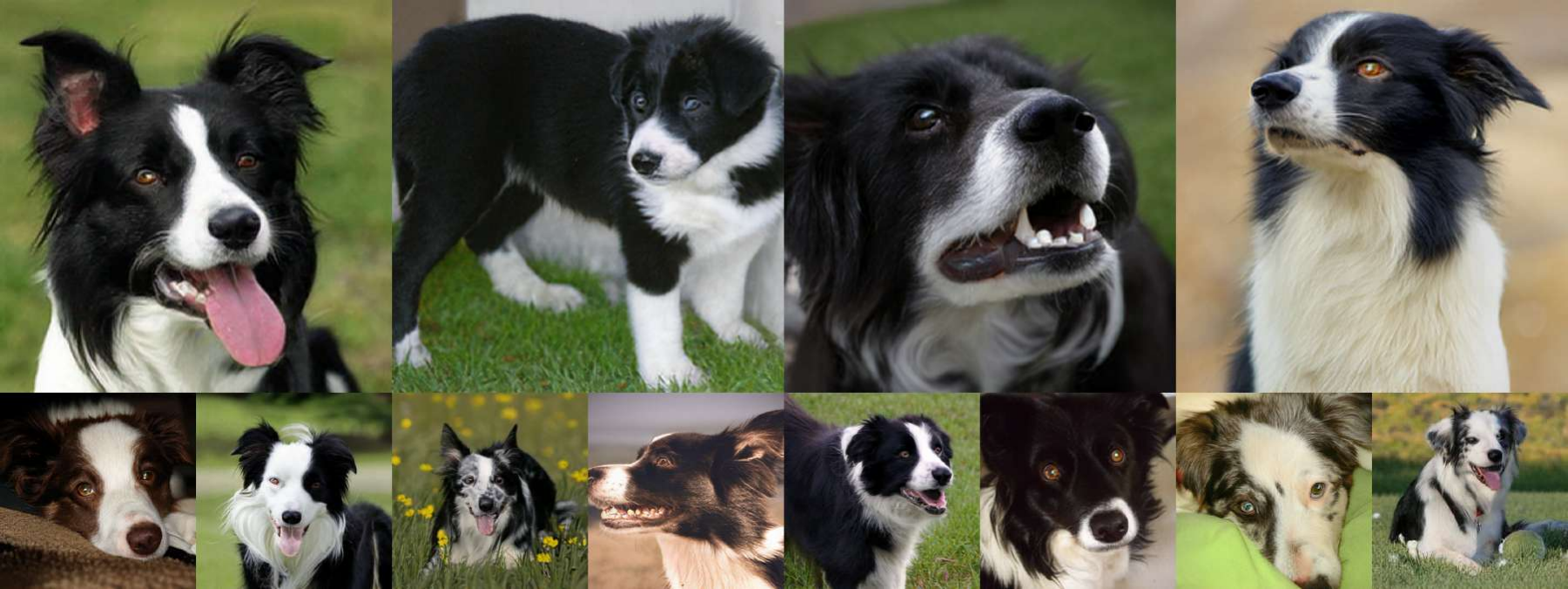}
  \caption{\textbf{Visualization of VFM-VAE + REG (640 epochs).}
  Generation uses CFG with $w=4.0$; class label is "Border collie" (232).}
  \label{fig:vfm-vae-reg-dog}
\end{figure*}

\begin{figure*}[ht]
  \centering
  \includegraphics[width=\textwidth]{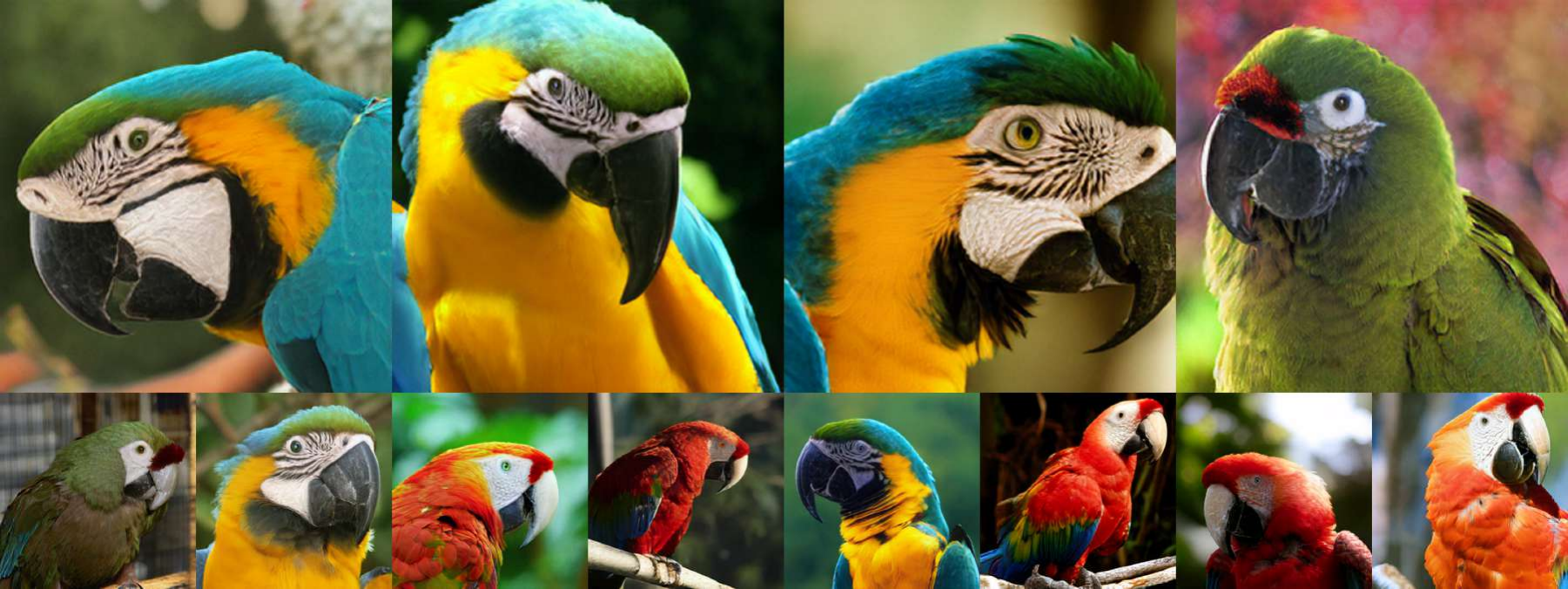}
  \caption{\textbf{Visualization VFM-VAE + REG (640 epochs).}
  Generation uses CFG with $w=4.0$; class label is "Macaw" (88).}
  \label{fig:vfm-vae-reg-macaw}
\end{figure*}

\begin{figure*}[ht]
  \centering
  \includegraphics[width=\textwidth]{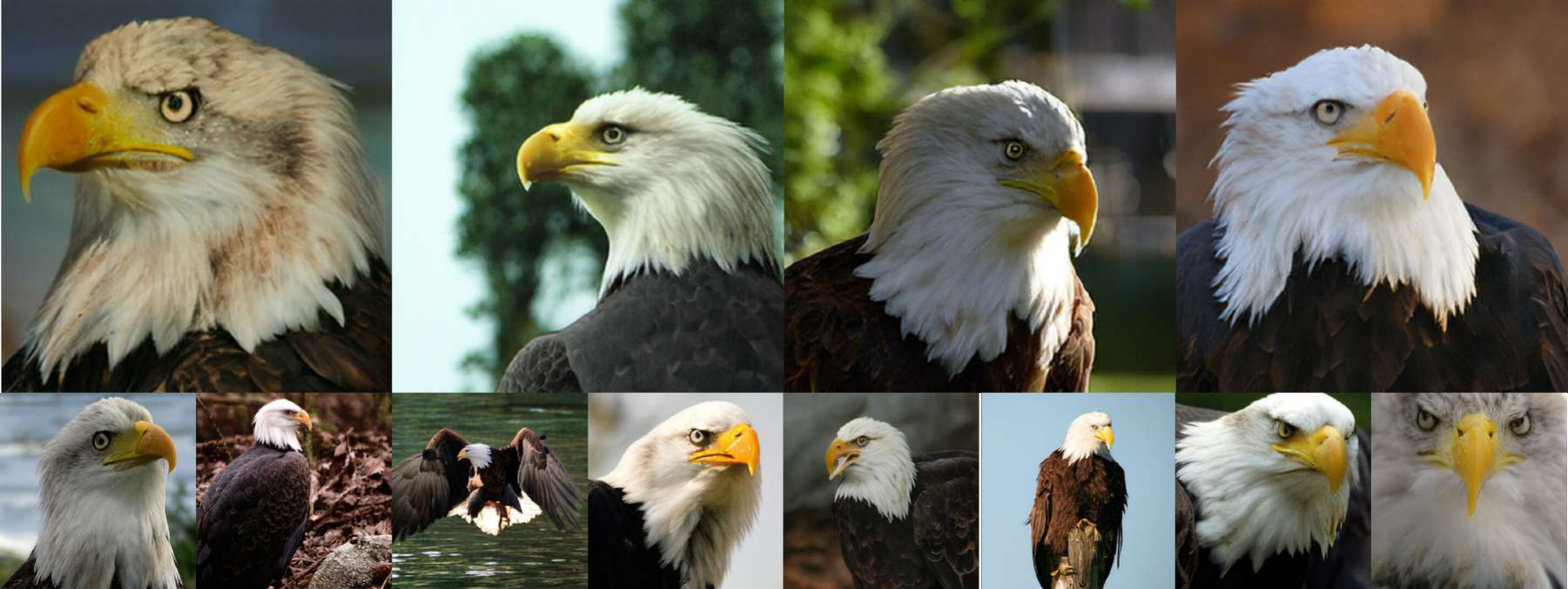}
  \caption{\textbf{Visualization of VFM-VAE + REG (640 epochs).}
  Generation uses CFG with $w=4.0$; class label is "Bald Eagle" (22).}
  \label{fig:vfm-vae-reg-eagle}
\end{figure*}

\begin{figure*}[ht]
  \centering
  \includegraphics[width=\textwidth]{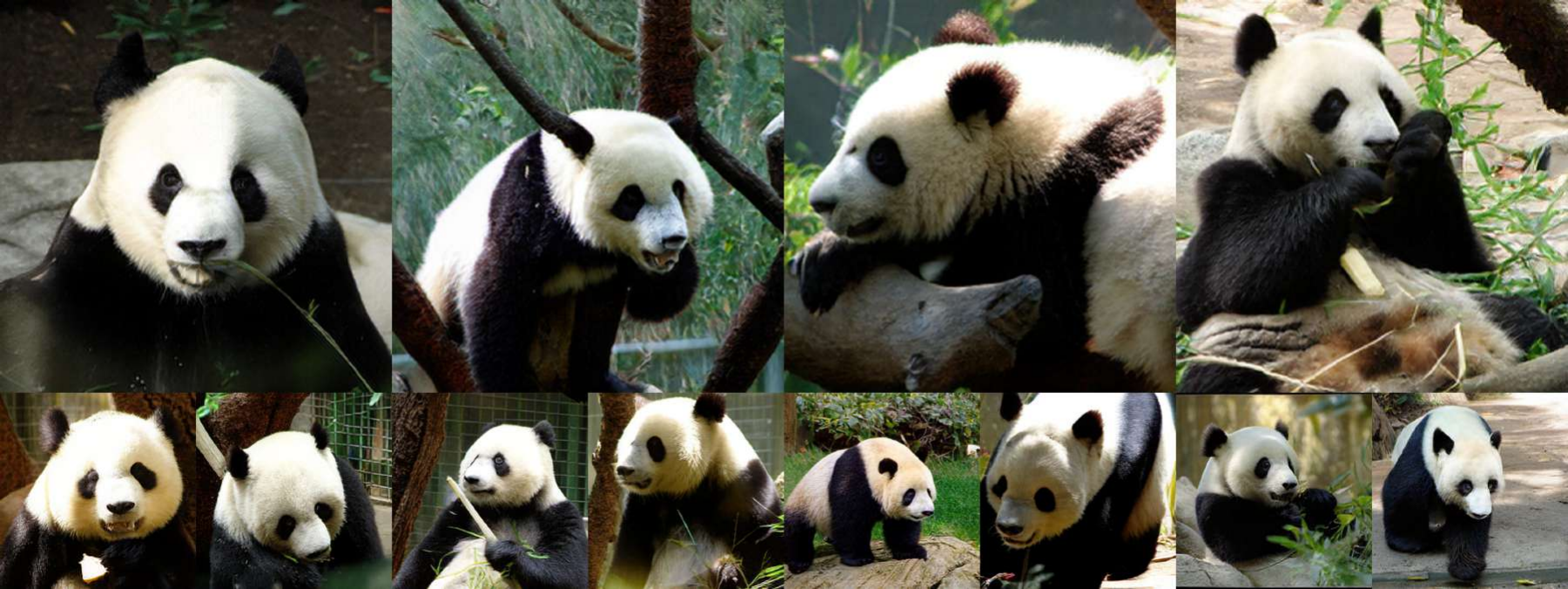}
  \caption{\textbf{Visualization of VFM-VAE + REG (640 epochs).}
  Generation uses CFG with $w=4.0$; class label is "Giant Panda" (388).}
  \label{fig:vfm-vae-reg-panda}
\end{figure*}

\begin{figure*}[ht]
  \centering
  \includegraphics[width=\textwidth]{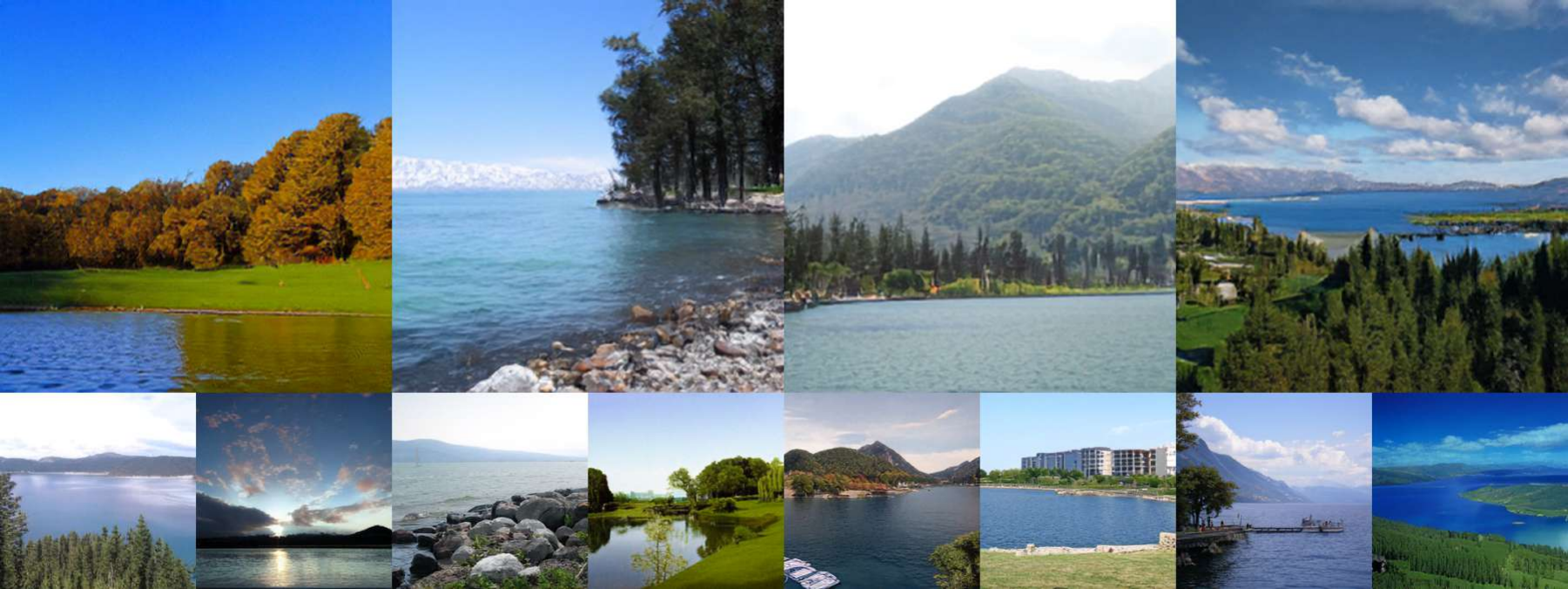}
  \caption{\textbf{Visualization of VFM-VAE + REG (640 epochs).}
  Generation uses CFG with $w=4.0$; class label is "Lakeside" (975).}
  \label{fig:vfm-vae-reg-lakeside}
\end{figure*}

\begin{figure*}[ht]
  \centering
  \includegraphics[width=\textwidth]{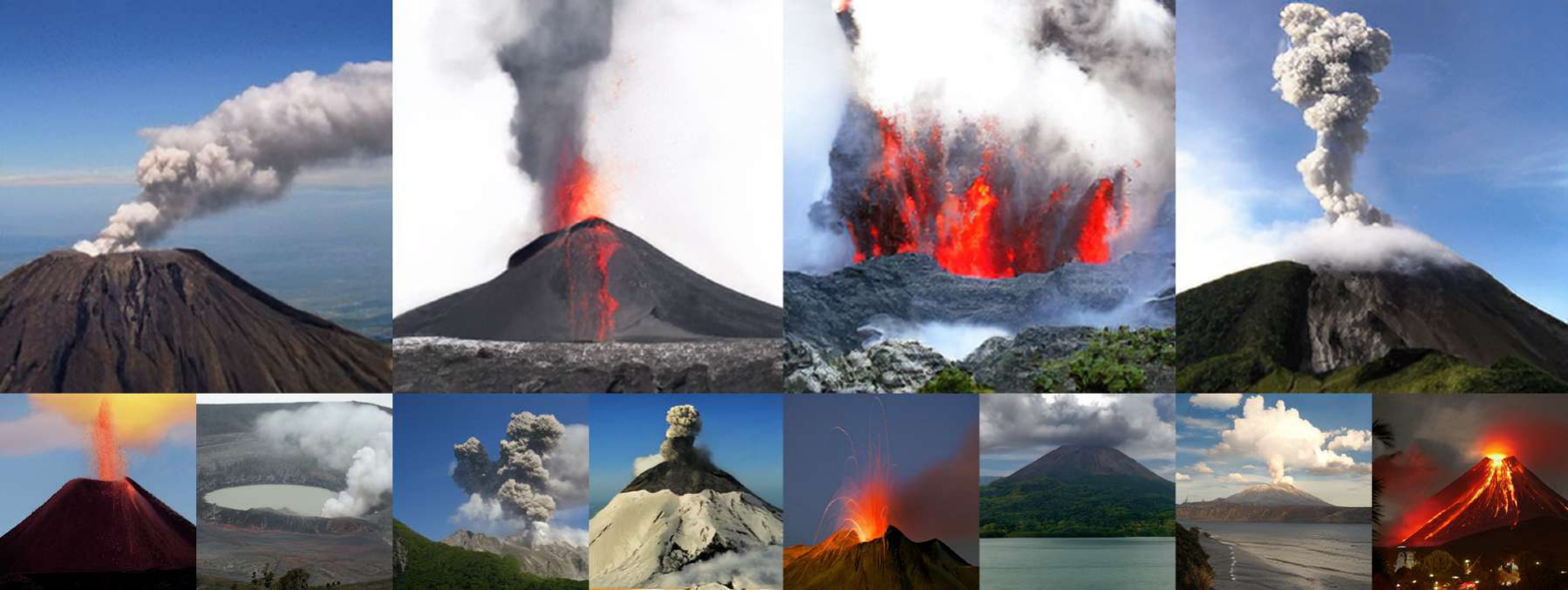}
  \caption{\textbf{Visualization of VFM-VAE + REG (640 epochs).}
  Generation uses CFG with $w=4.0$; class label is "Volcano" (980).}
  \label{fig:vfm-vae-reg-volcano}
\end{figure*}

We provide additional qualitative results: tokenizer reconstruction comparisons are shown in \cref{fig:vae_recon}, visualizations across the training stages of the generative model are presented in \cref{fig:generation}, and 256-resolution generation examples from VFM-VAE + REG are provided from \cref{fig:vfm-vae-reg-dog} to \cref{fig:vfm-vae-reg-volcano}.

